\documentclass[jair,twoside,11pt,theapa]{article}
\usepackage[papersize={8.5in,11in}]{geometry}
\usepackage{jair, theapa, rawfonts}
\usepackage{algorithm,algorithmic}
\usepackage{amscd,amsmath,amssymb}
\usepackage[acronym]{glossaries}
\usepackage{bbm}
\usepackage{float}
\usepackage{breakcites}
\usepackage{epsfig,epstopdf,floatflt,graphicx,subfigure, wrapfig}
\usepackage[acronym]{glossaries}
\usepackage{tikz, pb-diagram, pgfplots, pgfplotstable}

\usepackage{times}
\usepackage{ulem}
\usepackage{url}
\usepackage[makeroom]{cancel}

\renewcommand{\le}{\leqslant}
\renewcommand{\ge}{\geqslant}
\newcommand{\SUP}{\texttt{SUP}}
\newcommand{\SMPL}{\texttt{PMS$_2$L}}
\newcommand{\SVM}{\texttt{SVM}}

\newcommand{\MLP}{\texttt{McLP}}
\newcommand{\MTSVM}{\texttt{McTSVM}}
\newcommand{\SSGM}{\texttt{S$_2$GM}}

\newcommand{\EE}{\mathbb{E}}

\newcommand{\PP}{\mathbb{P}}
\newcommand{\RR}{\mathbb{R}}

\newcommand{\A}{\mathcal{A}}

\newcommand{\C}{\mathcal{C}}
\newcommand{\D}{\mathcal{D}}

\newcommand{\F}{\mathcal{F}}

\newcommand{\HHH}{\mathcal{H}}

\newcommand{\X}{\mathcal{X}}
\newcommand{\Y}{\mathcal{Y}}

\newcommand{\Inp}{\X}
\newcommand{\Outp}{\Y}
\makeatletter
\newcommand{\iid}[1]{\mathbin{\overset{#1}{\kern\z@\sim}}}%
\newsavebox{\mybox}\newsavebox{\mysim}
\makeatother
\newcommand{\lda}{{S_\ell}}

\newcommand{\se}{s_*}
\newcommand{\te}{t_*}
\newcommand{\ke}{k_*}

\newcommand{\neta}{n_\eta}
\newcommand{\ueta}{u_\eta}
\newcommand{\ulda}{{S_u}}

\newcommand{\obs}{\mathbf{x}}
\newcommand{\Indicator}[1]{\mathbbm{1}_{#1}}

\newcommand{\fR}{\mathfrak{R}}

\newtheorem{thrm}{Theorem}

\newtheorem{ddf}[thrm]{Definition}
\newtheorem{lemma}[thrm]{Lemma}

\newtheorem{cor}[thrm]{Corollary}

\newenvironment{Proof}{\par\noindent{\bf Proof.}}{\hfill$\square$}
\DeclareMathOperator*{\argmax}{arg\,max}

\newtheorem{innercustomgeneric}{\customgenericname}
\providecommand{\customgenericname}{}
\newcommand{\newcustomtheorem}[2]{%
  \newenvironment{#1}[1]
  {%
   \renewcommand\customgenericname{#2}%
   \renewcommand\theinnercustomgeneric{##1}%
   \innercustomgeneric
  }
  {\endinnercustomgeneric}
}

\newcustomtheorem{customthm}{Theorem}
\newcustomtheorem{customlemma}{Lemma}

\pgfplotsset{compat=newest}

\newcommand{\fleche}{{\tiny \downarrow}}

\ShortHeadings{Rademacher Complexity Bounds for a Penalized Multiclass Semi-Supervised Algorithm}
{Maximov, Amini, \& Harchaoui}
\firstpageno{1}

\begin{document}

\title{Rademacher Complexity Bounds for a Penalized Multiclass Semi-Supervised Algorithm}

\author{\name Yury Maximov \email yury@lanl.gov \\
      \addr Los Alamos National Laboratory \\
      Theortical Division T-4 and Center for Nonlinear Studies\\
      Los Alamos 87545 NM USA\\
      \AND
       \addr Skolkovo Institute of Science and Technology \\ 
         Skolkovo Innovation Center, Building 3-461 \\
         Moscow  143026 Russia \\
        \AND
       \name Massih-Reza Amini \email massih-reza.amini@imag.fr \\
             \addr Univ. Grenoble Alpes, CNRS, Grenoble INP, LIG \\
           F-38000 Grenoble Francee \\
       \AND
       \name Zaid Harchaoui \email zaid@uw.edu \\
       \addr University of Washington, Department of Statistics, \\
          Box 354322, Seattle, WA 98195-4322 USA 
       }


\maketitle

\begin{abstract}
We propose Rademacher complexity bounds for multiclass classifiers trained with a two-step semi-supervised model. In the first step, the algorithm partitions the partially labeled data and then identifies dense clusters containing $\kappa$ predominant classes using the labeled training examples such that the proportion of their non-predominant classes is below a fixed threshold stands for clustering consistency. In the second step,  a classifier is trained by minimizing a margin empirical loss over the labeled training set and a penalization term measuring the disability of the learner to predict the $\kappa$ predominant classes of the identified clusters. The resulting data-dependent generalization error bound involves the margin distribution of the classifier, the stability of the clustering technique used in the first step and Rademacher complexity terms corresponding to partially labeled training data. Our theoretical result exhibit convergence rates extending those proposed in the literature for the binary case, and experimental results on different multiclass classification problems show empirical evidence that supports the theory.
\end{abstract}

\section{Introduction}
\label{Introduction}

    Learning with partially labeled data, or Semi-supervised learning (SSL), has been an active field of study in the ML community these past twenty years. In this case, labeled examples are usually supposed to be very few leading to an inefficient supervised model, while unlabeled training examples contain valuable information on the prediction problem at hand which exploitation may lead to a performant prediction function. For this scenario, we assume available a set of labeled training examples $\lda=\left(\obs_i,y_i\right)_{1\leq i\leq n}\in(\Inp \times \Outp)^n$ drawn i.i.d. with respect to a fixed, but unknown, probability distribution $\D$ over $\Inp \times \Outp$ and a set of unlabeled training examples $\ulda=(\obs_{n+i})_{1\leq i\leq u}\in\Inp^u$ supposed to be drawn from the marginal distribution, $\D_{\X}$, over the domain $\Inp$. If $\ulda$ is empty, then the problem is cast into the supervised learning framework. The other extreme case corresponds to the situation where $\lda$ is empty and for which the problem reduces to unsupervised learning.

The issue of learnability with partially labeled data was studied under three related yet different hypotheses of \textit{smoothness assumption}, \textit{cluster assumption}, and \textit{low density separation} \cite{chapelle2006semi,zhu05survey} and many advances have been made on both algorithmic and theoretical front under these settings. 

Although classification problems, for which the design of SSL techniques is appealing, are multiclass in nature, the majority of theoretical results for semi-supervised learning has mainly considered the binary case \cite{Kaeaeriaeinen2005,Leskes2005,AminiLavioletteUsunier2008,El-YanivPechyony2009,BalcanBlum2010,UrnerShalev-ShwartzBen-David2011}. In this paper, we tackle the learning ability of multiclass classifiers trained on partially labeled data by first identifying dense clusters covering labeled and unlabeled examples and then minimizing an objective composed of the margin empirical loss of the classifier over the labeled training set, and also a penalization term measuring the disability of the learner to predict the predominant classes of dense clusters.

Our main result is a data-dependent generalization error bound for classifiers trained under this setting and which exhibits a complexity term depending on the effectiveness of the clustering technique to find homogenous regions of examples belonging to each class, the margin distribution of the classifiers and the Rademacher complexities of the class of functions in use defined for labeled and unlabeled data. The convergence rates deduced from the bound extends those proposed in the literature for the binary case, further experiments carried out on text and image classification problems, show that the proposed approach yields improved classification performance compared to extensions of state-of-the-art SSL algorithms to the multiclass classification case.

In the following section, we first define our framework, then the learning task we address. Section \ref{sec:theory} presents the Rademacher generalization bound for a classifier trained with the proposed algorithm. Section \ref{sec:discussion} positions our theoretical findings with respect to the state-of-the-art, and finally, section \ref{sec:exps} details experimental results that support this approach.

%
%
\begin{table}[t!]
\center
\caption{Notations}
\label{tab:notations}
\begin{tabular}{ll}
\hline
$\X\subseteq\mathbb{R}^d$ &  Input space,\\
$\Y=\{1,\ldots,K\}$ &  Output space,\\
$K$ (resp. $G$) &  Number of classes (resp. clusters),\\
$\lda$ (resp. $\ulda$) & The set of labeled (resp. unlabeled) training examples of size $n$ (resp. $u$),\\
$\A_Z:\X\rightarrow\{1,\ldots,G\}$ &  A clustering algorithm, $\A$, trained on the set $Z$,\\
$\Delta_n(\A_Z,\A_{Z'},\tilde Z)$ & Distance between two clusterings $\A_Z$ and $\A_{Z'}$ estimated over $\tilde Z$ (Eq. \ref{eq:Distance}),\\
$\Pi_{\ulda}$ & Partition of the unlabeled set obtained by $\A_\ulda$,\\
$\Pi^\star$ & Limit clustering of the input space obtained by $\A^\star$, a particular instantiation of  $\A$,\\
$\C_\kappa(\eta)$ &  The set of $\kappa$-uniformly bounded  clusters (Eq. \ref{eq:EtaClusters}),\\
$\Y_{\kappa}(\C)$ &  $\kappa$ most predominant classes found in cluster $\C$,\\
$m_h(\obs,y)$ & The margin of an example $(\obs,y)$ over the whole set $\Outp$ (Eq. \ref{eq:Margin}), \\
$m'_h(\obs, \Outp_\kappa(\C))$ & The margin of an unlabeled example taken with respect to $\Y_{\kappa}(\C)$ (Eq. \ref{eq:MarginUnl}),\\
$\mu_h(\obs)=\displaystyle{\mathop{\argmax}_{y\in\Y}}~ h(\obs,y)$ & The class prediction of  $h\in\mathbb{R}^{\mathcal{X}\times \mathcal{Y}}$ for an example $\obs$,\\
$\hat R_\rho(h)$ & Penalized empirical loss (Eq. \ref{eq:PEL}),\\
$\Omega_\rho(h,\C_\kappa(\eta))$ & Penalization term in $\hat R_\rho(h)$ estimated over $\C_\kappa(\eta)$ (Eq. \ref{eq:Pen}),\\
${\widehat R}_\rho(h, \C_j)$ & Empirical risk defined over a single cluster $\C_j \in \C_\kappa(\eta)$ (Eq. \ref{eq:EmpLossClust}).\\
\hline
\end{tabular}
\end{table}

%
%

\section{Penalized based semi-supervised multiclass classification}

We are interested in the study of multiclass classification problems where the output space is $\Outp~=~\{1,\ldots,K\}$, with $K > 2$.
The semi-supervised multiclass classification algorithm that we consider is tailored under the cluster assumption and operates in two steps  depicted in the following sections. 

%
%
\subsection{Partitioning of data and identifying $\kappa$-uniformly bounded clusters with level $\eta$}

The first step consists in partitioning the unlabeled training observations, into $G>0$ separate clusters with a clustering algorithm $\A$ trained on $\ulda$, denoted by $\Pi_{\ulda}$.

Clusters of $\Pi_{\ulda}$  that are well covered by classes in the labeled training set are then kept for learning the classifier (Section \ref{sec:LrnObj}).  Formally, for a fixed $\kappa\in\{1,\ldots,K\}$, let $\Outp_\kappa(\C)$ be the $\kappa$ most predominant classes of $\Outp$ present in  cluster $\C\in\Pi_{\ulda}$. We then define $\kappa$-uniformly bounded clusters with level $\eta$, $\C_\kappa(\eta)$, the set of clusters within $\Pi_\ulda$ that are covered by their $\kappa$ most predominant classes such that the proportion of  other classes within $\C$ not belonging to $\Outp_\kappa(\C)$ is less than $\eta/G$~:
\begin{equation} 
\label{eq:EtaClusters}
\C_\kappa(\eta)= \biggl\{\C\in\Pi_{\ulda} :  P_n((\obs,y)\in 
\C \wedge 
y\in \mathcal \Outp\setminus \Outp_\kappa(\C))\leq \frac{\eta}{G}\biggr\}.
\end{equation}

Where  $P_n$ the uniform probability distribution over $\lda$; defined for any subset $B\subseteq\lda$, as $P_n(B)=\frac{1}{n}\text{card}(B)$. 

%
%
\subsection{Learning objective}
\label{sec:LrnObj}
In the second step, we address a learning problem that is to find, in a hypothesis set $\HHH \subseteq \mathbb{R}^{\mathcal{X}\times \mathcal{Y}}$, a scoring function $h\in \HHH$ with low risk: 
\begin{equation}
\label{eq:GenError}
R(h)=\EE_{(\obs,y)\sim\D} \left[\Indicator{m_h(\obs,y)\le 0}\right],
\end{equation}
where $\Indicator{\pi}$ is the indicator function and $m_h(\obs,y)$ is the margin of the function $h$ at an example $(\obs,y)$ \cite{koltchinskii2002empirical}:
\begin{equation}
\label{eq:Margin}
m_h(\obs,y) = h(\obs,y) - \max\limits_{y'\in\Outp\setminus\{y\}} h(\obs,y').
\end{equation}
This is achieved by minimizing a penalized empirical loss, defined for a given $\rho>0$~:  
\begin{equation}
\label{eq:PEL}
{\widehat R}_\rho(h)={\widehat R}_\rho(h,\lda)+\Omega_\rho(h,\C_\kappa(\eta)),
\end{equation} 
composed of an empirical margin loss of $h\in\HHH$ on a labeled training set $\lda$,  
\begin{equation}
\label{eq:EmpUnl}
{\widehat R}_\rho(h,\lda) =\frac{1}{n}\sum_{(\obs, y)\in\lda} \Phi_\rho(m_h(\obs,y)),
\end{equation}
and a penalization term that reflects the ability of the hypothesis $h\in\HHH$ to identify the $\kappa$ most predominant classes within the disjoint clusters of $\C_\kappa(\eta)$;  \begin{equation}
\label{eq:Pen}
\Omega_\rho(h, \C_\kappa(\eta)) = \frac{1}{u} \sum_{\C\in\C_\kappa(\eta)}\sum_{\obs\in \C}\Phi_\rho(m'_h(\obs, \Outp_\kappa(\C))),
\end{equation}
 where $m'_h(\obs, \Outp_\kappa(\C))$ is the margin of an unlabeled example taken with respect to the set of $\kappa$ predominant classes, $\Y_\kappa(\C)$ :
\begin{equation}
\label{eq:MarginUnl}
m'_h(\obs, \Y_\kappa(\C))= \max\limits_{y\in \Y_\kappa(\C)} h(\obs,y) -  \max\limits_{y\in \Y\setminus\Y_\kappa(\C)} h(\obs,y), \; \obs\in\C\subset \C_\kappa(\eta),
\end{equation}
and,  $\Phi_\rho:\mathbb{R}\rightarrow [0,1]$ is the $\rho$-margin loss defined as  \cite{koltchinskii2002empirical} :
\begin{equation}
\label{eq:RhoMArgin}
\forall z\in\mathbb{R}, \Phi_\rho(z) = 
\begin{cases}
0 & \text{if } \rho \ge z,\\ 
1 - z/\rho & \text{if } 0 < z < \rho, \\
1 & \text{if } z\le 0.
\end{cases}
\end{equation}
%
%
\begin{figure*}[t!]
\renewcommand{\figurename}{Algorithm}
\renewcommand\thefigure{1}
\hrule
  \caption{Pseudo-code of the {\SMPL} algorithm}
  \vspace{2mm}
\hrule
   \label{alg:smplpca}
\begin{algorithmic}
 \vspace{2mm}  \STATE {\bfseries Input:} Labeled data set $\lda=\left(\obs_i,y_i\right)_{1\leq i\leq n} \subseteq (\X\times\Y)^n$;
   \STATE \hspace{1cm} Unlabeled data set $\ulda=(\obs_{n+i})_{1\leq i\leq u} \subseteq \X^u$;
     \STATE \hspace{1cm} Hypothesis space $\HHH$;
    \STATE \hspace{1cm} $G$ the number of clusters, $\A_{\ulda}:\Inp\rightarrow\{1,\ldots,G\}$ the clustering algorithm found on $\ulda$,  $\kappa\in\mathbb{N}^*$, and $\eta>0$;
   \STATE {\bfseries Stage~1: }
   Using the labeled examples, $\lda$, identify the $\kappa$-bounded clusters in $\Pi_{\ulda}$ with level $\eta, \C_\kappa(\eta)$;  \textcolor{gray}{// in accordance with Eq. \eqref{eq:EtaClusters} }
   \STATE {\bfseries Stage~2: } Find a hypothesis $h^*\in\HHH$ that minimizes the penalized objective function (Eq. \ref{eq:PEL}) :
   \[
   h^*=\mathop{\text{argmin}}_{h\in\HHH} \widehat{R}_{\rho}(h)
   \]
   \STATE {\bfseries Output:} $h^*$
\end{algorithmic}
\hrule
\end{figure*}
Table \ref{tab:notations} summarizes notations used throughout the paper and the pseudo-code of the proposed 2-step approach, referred to as Penalized Multiclass Semi-Supervised Learning ({\SMPL}) in the following, is given in algorithm \ref{alg:smplpca}. 

\smallskip

The algorithm shares similarities with algorithms proposed in~\cite{AminiTruongGoutte2008,UrnerShalev-ShwartzBen-David2011}, where the $k$-NN technique was used to increase the size of the labeled training data by pseudo-labeling unlabeled examples that are in the nearest neighborhood of labeled examples, for binary classification and bipartite ranking. In \cite{Rigollet2007}, another two-step semi-supervised procedure is proposed where in the first stage a clustering of the feature space derived from the unlabeled data is produced and then each unlabeled observation, in a given cluster is assigned the same class label than the majority of labeled examples belonging to that class within the cluster.

In the present work we tackle a more general situation by considering multiclass classification problems and by relaxing the pseudo-labeling part which may be too aggressive in the multiclass case. Our analysis is based on the ability of a clustering technique to capture the structure of the data, and the ability of the classifier to identify predominant classes in $\kappa$-uniformly bounded clusters, leading to a multiclass definition of the cluster assumption which states that penalization over $\kappa$-uniformly bounded clusters with a bounded confident level $\eta$ helps learning.

%
%

\section{Theoretical study}
\label{sec:theory}

We now analyze how the use of unlabeled training data can improve generalization performance in some cases. Essentially, the trade-off is that clustering offers additional knowledge on the problem, therefore potentially helps learning, but can also be of lower quality, which may degrade it.

%
%
\subsection{Stable clustering with the bounded difference property}

Before, let us first introduce notations that are used in the statement of the following results. We consider a hard clustering algorithm $\A_Z$ defined as a function found over a finite sample $Z$. 

Our analyzes are based on a notion of stability of the clustering algorithm $\A_.$;   measured as the average number of examples in a given set $\tilde Z$ of size $n$ that are in the exclusive disjunction of clusters  (present in one and absent from the other) found by $\A_.$ over two sets $Z$ and $Z'$, and defined as~:
\begin{equation}
\label{eq:Distance}
  \Delta_n\left(\A_{Z}, \A_{Z'}, \tilde Z\right) = \mathop{\min}_{\pi}\left[\frac{1}{n} \sum\limits_{\obs\in \tilde Z} \Indicator{\A_Z(\obs)\neq \pi(\A_{Z'}(\obs))}\right],
\end{equation}
where $\pi:\{1,\ldots,G\}\rightarrow \{1,\ldots,G\}$ is a permutation. It is straightforward to show that $\Delta_n$ defines a true metric, sometimes referred to as the minimal matching distance \cite{von2010clustering},  on the space of clusterings (see Th. \ref{thm:DeltaDist} in the Appendix). Hence, the clustering algorithm $\A_.$ is said to obey the bounded difference property, if and only if  for any i.i.d. samples $Z$, $Z' \sim {\cal D}_{\cal X}^{|Z|}$ differing in exactly one observation, and for any i.i.d. sample  $\tilde Z\sim {\cal D}_{\cal X}^n$ of size $n$, there exists a universal constant~$L$ such that~:
\begin{gather}\label{eq:bounded-diff-prop}
\Delta(\A_{Z}, \A_{Z'}) = \EE_{\tilde Z \sim \D_\X^n}\left[ \Delta_n\left(\A_{Z}, \A_{Z'}, \tilde Z\right)\right]\le \frac{L}{|Z|}.
\end{gather}

For some clustering algorithms such as $k$-means or $k$-hyperplane clustering, it has been shown that the bounded difference property is tightly related to their (in)stability.  We refer to \cite{von2010clustering,LuxburgBousquetBelkin2004,RakhlinCaponnetto2006,ThiagarajanRamamurthySpanias2011} and a number of references therein for the algorithmic details as well as various notions of clustering instability, and to \cite{ShamirTishby2007} for the relation between bounded differences property, stability and model selection. Furthermore, in the case where a clustering algorithm $\A$ obeys the bounded difference property; it is said to be stable if  for any distribution $\D_\X$ over $\X$ there exists a unique limit clustering of the input space $\Pi^\star$, obtained by a particular instantiation of the algorithm  denoted by $\A^\star$, such that for any $Z$ drawn i.i.d. from $\D_\X$ and for any sample $\tilde Z$ of size $n$ drawn i.i.d. from the same distribution we have~: 
\begin{gather}\label{eq:consistency-prop}
\EE_{Z\sim \D_\Inp^{|Z|}} \left[\Delta(\A_Z, \A^\star)\right] \le \frac{L}{|Z|}.
\end{gather}
In this case, it is possible to (tightly) upper-bound the distance between $\A^\star$ and the algorithm $\A$ trained on any unlabeled training set $\ulda$, estimated over the labeled training set $\lda$: $\Delta_n(\A_{\ulda}, \A^\star, \lda)$, as it is stated in the following Lemma.

%
%
\begin{lemma}
\label{lem:class-stability}
Let $\lda=\left(\obs_i,y_i\right)_{1\leq i\leq n}$ and $\ulda=(\obs_{n+i})_{1\leq i\leq u}$ be a labeled and an unlabeled training sets drawn i.i.d. according respectively to a probability distribution $\D$ over $\Inp \times \Outp$,  and its marginal $\D_\Inp$. For any $1>\delta>0$ and any stable clustering algorithm $\A$ that obeys the bounded differences property with constant $L>0$, the average number of examples in $\lda$ that are in the exclusive disjunction of clusters found by the clustering algorithm $\A$ on $\ulda$ and by $\A^\star$ is upper-bounded  with probability at least $1-\delta$ as follows~: 
\begin{equation}
    \label{eq:DeltaBnd}
\Delta_n(\A_{\ulda}, \A^\star, \lda) \le \frac{L}{u} + L\sqrt{\frac{\log \frac{2}{\delta}}{2u}} + \sqrt{\frac{ \log \frac{2}{\delta}}{2n}}.
\end{equation}

\end{lemma}

The proof is given in Appendix \ref{sec:Proofs}. This result suggests that for any labeled and unlabeled training data, if a clustering algorithm obeys the bounded differences property and that it is stable, then with high probability, $\Pi_\ulda$ covers as well the labeled training data as the limit partition $\Pi^\star$ (i.e. most of the labeled examples would more likely be present in  the intersection $\Pi_\ulda\cap\Pi^\star$).
%
%




%
%

\subsection{Semi-supervised Data-dependent bounds}

Based on the previous lemma, we can define situations where the Empirical Risk Minimization principle of algorithm {\SMPL} becomes consistent. This result is stated in Theorem \eqref{thrm:main-multi-simple} which provides bounds on the generalization error of a multiclass classifier trained with the penalized empirical loss defined above (Eq. \ref{eq:PEL}).

The notion of function class capacity used in the bounds, is the labeled and unlabeled Rademacher complexities of the function class $\F_\HHH= \{f:\obs\mapsto h(\obs,y): y\in\Outp, h\in\HHH\}$, defined respectively as: 
\begin{align*}
\fR_n^*(\F_\HHH) &=\!\!\!\sum\limits_{\C \in \C_{\kappa}(\eta)} \EE_{\sigma} \sup\limits_{f\in\F_\HHH} \frac{2}{n}
\sum_{\obs_i\in\lda\cap\C}\sigma_i f(\obs_i)
, \\
\fR_u^*(\F_\HHH) &=\!\!\!\sum\limits_{\C \in \C_{\kappa}(\eta)} \EE_{\sigma} \sup\limits_{f\in\F_\HHH} \frac{2}{u}
\sum_{\obs_i\in\ulda\cap\C}\sigma_i f(\obs_i)
,\\
\fR_n(\F_\HHH) &=\EE_{\sigma} \sup\limits_{f\in\F_\HHH} \frac{2}{n}
\sum_{\obs_i\in\lda\setminus\C_{\kappa}(\eta)}\sigma_i f(\obs_i)
\end{align*}

where $\sigma_i$'s, called Rademacher variables, are independent uniform random variables  taking values in $\{-1, +1\}$; i.e. $\forall i, \PP(\sigma_i=-1)=\PP(\sigma_i=+1)=\frac{1}{2}.$

\smallskip

\noindent The proof of the theorem is based on the following Lemma that provides generalization bounds over the true risk of any classifier $h$, found by algorithm {\SMPL} and estimated within a single confident cluster; $\C_j\in \C_\kappa(\eta)\subseteq \Pi_\ulda$~:

\begin{equation}
\label{eq:TrLossClust}
R(h, \C_j)=\EE [\mu_h(\obs) \neq y \wedge \obs\in \C_j],
\end{equation}
with respect to the estimated empirical risk~:
\begin{equation}
\label{eq:EmpLossClust}
{\widehat R}_\rho(h, \C_j)=\frac{1}{n}\sum_{(\obs, y)\in\lda\cap \C_j} \Phi_\rho(m_h(\obs,y))+\frac{1}{u} \sum_{\obs\in\ulda\cap\C_j}\Phi_\rho(m'_h(\obs, \Outp_\kappa(\C_j))).
\end{equation}

\begin{lemma}
\label{lem:rademacher}
Let $\HHH \subseteq \mathbb{R}^{\Inp\times \Outp}$ be a hypothesis set where $\Outp=\{1,\ldots,K\}$, and let $\lda=\left(\obs_i,y_i\right)_{1\leq i\leq n}$ and $\ulda=(\obs_{n+i})_{1\leq i\leq u}$ be two sets of labeled and unlabeled training data, drawn i.i.d. respectively according to a probability distribution over $\Inp \times \Outp$ and a marginal distribution $\D_{\X}$. Fix $\rho>0$, $\kappa\in\{1,\ldots,K\}$ 
then for any $1>\delta > 0$, the following multiclass classification generalization error bound holds with probability at least $1-\delta$ 
for all $h\in \HHH$ learned by algorithm \ref{alg:smplpca} over a single $\kappa$-uniformly bounded cluster $\C_j\in \C_\kappa(\eta)$ derived from $\ulda$ by a clustering algorithm $\A_\ulda$ that partitions the input space into $G$ clusters~:
    \begin{align*}
     R(h, \C_j) & \leq {\widehat R}_{\rho}(h, \C_j) + \frac{\eta}{G} + \frac{2\kappa}{\rho}\fR_{n,j}^*(\F_\HHH) + \frac{2K}{\rho} \fR^*_{u,j}(\F_\HHH)  \\ 
     &\quad + 5\sqrt{\frac{\kappa \neta(j)\log \frac{16K}{\delta}}{2n^2}} + 5\sqrt{\frac{\kappa \ueta(j)\log \frac{16K}{\delta}}{2u^2}} + \frac{7\log \frac{8}{\delta}}{3(n-1)}  + \frac{7\log \frac{8}{\delta}}{3(u-1)},
   \end{align*}
\noindent 
where $\neta(j) = |\lda \cap \C_j|$, $\ueta(j) = |\ulda \cap \C_j|$, $\fR_{n,j}^* = \EE_{\sigma} \sup\limits_{f\in\F_\HHH} \frac{2}{n}
\sum_{\obs_i\in\lda\cap\C_j}\sigma_i f(\obs_i)
$, and  $\fR_{u,j}^* = \EE_{\sigma} \sup\limits_{f\in\F_\HHH} \frac{2}{u}
\sum_{\obs_i\in\ulda\cap\C_j}\sigma_i f(\obs_i)
$.
\end{lemma}

\bigskip

The proof is provided in Appendix \ref{sec:Proofs}. From this result and Lemma \ref{lem:class-stability}, we can then derive a data-dependent generalization bound for any semi-supervised multiclass prediction function found by algorithm {\SMPL} as stated below.

\begin{thrm}
  \label{thrm:main-multi-simple}
Let $\HHH \subseteq \mathbb{R}^{\Inp\times \Outp}$ be a hypothesis set where $\Outp=\{1,\ldots,K\}$, and let $\lda=\left((\obs_i,y_i)\right)_{i=1}^n$ and $\ulda=(\obs_i)_ {i=n+1}^{n+u}$ be two sets of labeled and unlabeled training data, drawn i.i.d. respectively according to a probability distribution over $\Inp \times \Outp$ and a marginal distribution $\D_{\X}$. Fix $\rho>0$ and $\kappa\in\{1,\ldots,K\}$, and consider a clustering algorithm $\A$ that obeys the bounded difference property with constant $L$ and is stable. If the $\kappa$-uniformly bounded clusters found in $\Pi_\ulda$ are such that the confident level $\eta$ satisfies $\eta\leq \Delta_n(\A_\ulda,\A^\star,\lda)$,
then for any $1>\delta > 0$ and all $h\in \HHH$ found by the {\SMPL} algorithm using $\A_\ulda$, the following multiclass classification generalization error bound holds with probability at least $1-\delta$~:
   \begin{equation*}
     R(h) \hspace{-1mm}\leq \hspace{-1mm} {\widehat R}_{\rho}(h)   + \frac{L}{u} +\frac{2K}{\rho} (\fR^*_u(\F_\HHH)+ \fR_n(\F_\HHH)) + \frac{2\kappa}{\rho}\fR^*_n(\F_\HHH)+ \frac{7G\log\hspace{-.3mm}\frac{14G}{\delta}}{3\se} + \sqrt{\frac{\hspace{-.3mm}\log\hspace{-.3mm} \frac{14}{\delta}}{\te}}+9\sqrt{\hspace{-.3mm}\frac{\log \hspace{-.3mm}\frac{14KG}{\delta}}{v_*}},
   \end{equation*}
\noindent 
where 
$\frac{1}{\se}\doteq \left(\frac{2G}{n-1}+\frac{G}{u-1}\right), \frac{1}{\te} \doteq \frac{L^2}{u}+\frac{1}{n}, \frac{1}{v_*} \doteq \frac{G\kappa \ueta}{2u^2} + \frac{G \kappa\neta+K(n-\neta)}{2n^2}, \neta = |\lda \cap \C_\kappa(\eta)|$ and $\ueta = |\ulda \cap \C_\kappa(\eta)|$. 
\end{thrm}

\bigskip

The proof is provided in Appendix \ref{sec:Proofs}. This result implies that with stable clustering algorithms obeying the bounded differences property, if the proportion of other classes than $\kappa$-predominant ones in confident clusters is less than the number of labeled examples in the exclusive disjunction of limit clusters and those found using the unlabeled training data, then with the strategy defined in algorithm {\SMPL} we can expect to have interesting situations for learning prediction models as it is stated in the following corollary.

Consider kernel-based hypotheses with $\mathfrak{K}: \Inp\times\Inp\rightarrow \RR$ a PSD kernel and $\Phi:\Inp\rightarrow\mathbb{H}$ its associated feature mapping function, defined as :
\[
\mathcal{H}_B=\left\{(\obs,y)\in\Inp\times\Outp \mapsto \langle \Phi(\obs),\mathbf{w}_y \rangle \mid \mathbf{W}=(\mathbf{w}_1, \ldots, \mathbf{w}_K), \left\| \mathbf{W}\right\|_{\mathbb{H},2} \leq B\right\}.
\]

Where  $\left\| \mathbf{W}\right\|_{\mathbb{H},2}$ is  the Frobenius norm of the parameter matrix for a linear
kernel, or the $L_{\mathbb{H},2}$ group norm of $\mathbf{W}$,  defined as 
\[
\left\| \mathbf{W}\right\|_{\mathbb{H},2}=\sqrt{\sum_{k=1}^K \left\| \mathbf{w}_k\right\|_{\mathbb{H}}^2}.
\]
In this case, we can derive the  following corollary from theorem \ref{thrm:main-multi-simple}~:

\begin{cor}
\label{thm:cor}
Let $\mathfrak{K}: \Inp\times\Inp\rightarrow \RR$ be a PSD kernel and  let $\Phi:\Inp\rightarrow\mathbb{H}$ be the associated feature mapping function. Assume that there exists $R>0$ such that $\mathfrak{K}(\obs,\obs)\leq R^2$ for all $\obs\in\Inp$. Then for any $1>\delta > 0$ and under the conditions and the definitions of theorem \ref{thrm:main-multi-simple}, the following multi-class classification error bound holds for all hypothesis $h\in\mathcal{H}_B$ learned  by the proposed algorithm over the set of $\kappa$-uniformly bounded set of clusters, $\C_\kappa(\eta)$, with probability at least $1-\delta$ :
     \begin{gather*}
 R(h) \leq {\widehat R}_{\rho}(h) + \frac{L}{u} + \frac{2}{\rho} RB\sqrt{\frac{3\ke^2}{\se}}+ \frac{7\log\frac{14G}{\delta}}{3\se} + \sqrt{\frac{\log\frac{14}{\delta}}{\te}} +  5\sqrt{\frac{3\log \frac{14KG}{\delta}}{v_*}} 
      \end{gather*}
\end{cor}
where $\frac{1}{\se}\doteq \left(\frac{2G}{n-1}+\frac{G}{u-1}\right), \frac{1}{\te} \doteq \frac{L^2}{u}+\frac{1}{n}, \frac{1}{v_*} \doteq \frac{G\kappa \ueta}{2u^2} + \frac{G \kappa\neta}{2n^2} + \frac{K(n-\neta)}{2n^2}$ and $\frac{\ke^2}{v_*} \doteq K^2 \frac{G \ueta}{u^2} + \kappa^2\frac{G \neta}{n^2} + K^2 \frac{n-\neta}{n^2}$.

\bigskip

\begin{Proof}
 From the proposition (8.1) in \cite{MohriRostamizadehTalwalkar2012}, and the Cauchy-Schwartz inequality $\left(\sum_{j=1}^G a_j b_j\right)^2 \le \left(\sum_{j=1}^G a_j^2\right) \left(\sum_{j=1}^G b_j^2\right)$ with $b_j = 1$ and $a_j = \sqrt{\ueta(j)}, \forall j$; the Rademacher complexity of the class of linear classifiers in the feature space can be bounded as~: 
  \[
  \fR^*_u(\F_\HHH)\le \sum\limits_{\C_j\in \C_\kappa(\eta)} \frac{2}{u} RB \sqrt{\ueta(j)} \le 2RB \sqrt{\frac{G\ueta}{u^2}},
  \]
  where $\ueta(j)$ in the number of unlabeled examples in $\eta$-confident cluster $\C_j$ and $\ueta=\sum_j \ueta(j)$ is the total number of unlabeled examples within a set of confident clusters $\C_\kappa(\eta)$.
  
  Similarly, if $\neta(j)$ is the number of unlabeled examples in  $\C_j\in \C_\kappa(\eta)$ we have~:
    \[
  \fR_n^*(\F_\HHH)\le \sum\limits_{\C_j \in \C_\kappa(\eta)} \frac{2}{n} RB \sqrt{\neta(j)} \le 2RB \sqrt{\frac{G\neta}{n^2}},
  \]
   and also  $\fR_n(\F_\HHH)\le 2RB \sqrt{\frac{n-\neta}{n^2}}$. Applying the Cauchy-Schwartz inequality again we finally get~:
    \[
    2K \fR_n(\F_\HHH) + 2\kappa\fR_n^*(\F_\HHH) + 2K\fR_u^*(\F_\HHH) \le 2RB  \sqrt{\frac{3\ke^2}{\se}}
    \]\end{Proof}

\bigskip

The non-empirical terms of this bound determine the convergence rate of the proposed penalized semi-supervised mutliclass algorithm, and hence following \cite[theorem 2.1, p.38]{vapnik2000nature}, gives insights on its consistency. These terms may be better explained using orders of 
magnitude~\cite{Knuth76}.  If we now consider the common situation in semi-supervised learning where $u\gg n$, and $\neta \approx n, \ueta\approx u$, and $\kappa=O(1)$, $L = O(1)$, $G = O(K)$ then   

\[
\frac{\ke^2}{v_*} \doteq K^2 \frac{G \ueta}{u^2} + \kappa^2\frac{G \neta}{n^2} + K^2 \frac{n-\neta}{n^2} = O\left(\frac{K^3}{u} + \frac{K}{n}\right),
\] 

\[\frac{1}{v_*} \doteq \frac{G\kappa \ueta}{u^2} + \frac{G \kappa\neta}{n^2} + \frac{K(n-\neta)}{n^2} = O\left(\frac{K}{u} + \frac{K}{n}\right),\]

and
\[
\frac{1}{\se}=O\left(\frac{K}{n}+\frac{K}{u}\right),~~ \frac{1}{\te} \doteq \frac{L^2}{u}+\frac{1}{n} = O\left(\frac{1}{u} + \frac{1}{n}\right).
\]

The convergence rate of the bound of corollary \ref{thm:cor} is of the order 
\begin{equation} 
\label{eq:convrate}
\tilde{O}\left(\sqrt{\frac{K}{n}}+K\sqrt{\frac{K}{u}}\right),
\end{equation}
where, for any real valued functions $f$ and $g$ the equality ;    $f(z) = \tilde O (g(z))$ holds,   if there exists a constant $\alpha > 0$ such that $f(z) = O(g(z)\log^\alpha g(z))$ \cite{Knuth76}. In the following section we present an overview of the related-work and show that in the case where the clustering technique $\mathcal A$ captures the true structure of the data, measured by the set of $\kappa$-uniformly bounded clusters with rate $\eta$, resulting in approximations above, then for linear  kernel-based hypotheses, the convergence rate \eqref{eq:convrate} is the direct extension of dimension-free convergence rates proposed in semi-supervised learning for the binary case.

As for the opposite case $n \gg u$ the pseudo-labeling step does not help to learning and even can make the bounds worse than at the supervised case. The same situation takes place when the number of classes is comparable to the number of objects and one can not clarify whether a cluster is consistent or not. 

Finally we would like to emphasize that our main target is the most practical case with $u\gg n$ and the number of classes comparable to the number of clusters.

%

\section{Related works and discussion}
\label{sec:discussion}

Semi-supervised learning (SSL) approaches exploit the geometry of data to learn a prediction function from partially labeled training sets \cite{seeger2000learning}. The three main SSL techniques; namely  graphical, generative and discriminant approaches, were mostly developed for the binary case and tailored under smoothness, low density separation and cluster assumptions \cite{zhu05survey,chapelle2006semi,AminiU15}. 

Graphical approaches construct an empirical graph where the nodes represent the training examples and the edges of the graph reflect the similarity between them. These approaches are mostly based on label spreading algorithms that propagate the class label of each labeled node to its neighbors \cite{ZhouBousquetLalEtAl2003,zhu02ssl}. Generative approaches naturally exploit the geometry of data by modelling their marginal distributions. These methods are developed under the cluster assumption  and use the Bayes rule to make decision. In the seminal work of \cite{CastelliCover1995} it is shown that, without extra assumptions relating marginal distribution and true distribution of labels, a sample of unlabeled data is of (almost) no help for learning purpose. Recent work from \cite{Ben-DavidLuPal2008} investigated further the limitations of semi-supervised learning and concluded that theoretical results for semi-supervised learning should be accompanied by an extra assumption on the true label distribution. 

Discriminant approaches directly find the decision boundary without making any assumptions on the marginal distribution of examples. The two most popular discriminant models are without doubts co-training \cite{BlumMitchell1998}  and Transductive SVMs \cite{vapnik2000nature}. The co-training algorithm supposes that each observation is produced by two sources of information and that each view-specific representation is rich enough to learn the parameters of the associated classifier in the case where there are enough labeled examples available. The two classifiers are first trained separately on the labeled data. A subset of unlabeled examples is then randomly drawn and pseudo-labeled by each of the classifiers. The estimated output by the first classifier becomes the desired output for the second classifier and reciprocally. Under this setting, \cite{Leskes2005} proposed a Rademacher complexity bound, where unlabeled data are used to decrease the disagreement between hypotheses from a class of functions $\HHH$ and proved that in some cases, the bound of the excess risk $|R(h)-\hat R(h,\lda)|$ for any $h\in\HHH$ is of the order $\tilde O\left(n^{-1/2}+u^{-1/2}\right)$. Another study in this line of research is \cite{TolstikhinZB15}. However, transductive learning tends to produce a prediction function for only a fixed number of unlabeled examples.  Transductive algorithms generally use the distribution of unsigned margins of unlabeled examples in order to guide the search of a prediction function and find the hyperplane in a feature space that separates the best labeled examples and that does not pass through high density regions. The notion of transductive Rademacher complexity was introduced in \cite{El-YanivPechyony2009}. In the best case, the excess risk bound proposed in this paper is of the order  $\tilde O\left(u\sqrt{\min(u,n)}/(n+u)\right)$.

\begin{table}[!b]
 \caption{\small Summary of the  convergence rates of dimension free bounds of excess risks for different SSL approaches.  }
\begin{center}
\begin{tabular}{l|l}
      \label{tab:known-rates}
 Order of convergence rates & Case; Reference \\
\hline\noalign{\smallskip}
 $\tilde{O}\left(\frac{u\sqrt{\min(u,n)}}{n+u}\right)$ & Binary; \cite{El-YanivPechyony2009}\\
\noalign{\smallskip}\hline\noalign{\smallskip}
 $\tilde{O}\left(\frac{1}{n} + \frac{1}{\sqrt{u}}\right)$ & Binary;\cite{BalcanBlum2010}\\
\noalign{\smallskip}\hline\noalign{\smallskip}
 $\tilde{O}\left(\frac{1}{\sqrt{n}} + \frac{1}{\sqrt{u}}\right)$ & Binary; \cite{Leskes2005} \\
\noalign{\smallskip}\hline\noalign{\smallskip}
 $\tilde{O}\left(\frac{1}{\sqrt{n}} + \frac{1}{\sqrt{u}}\right)$& Binary; \cite{Kaeaeriaeinen2005}\\
\noalign{\smallskip}\hline\noalign{\smallskip}
$\tilde{O}\left(\frac{\sqrt{K}}{\sqrt{n}}+\frac{K^{3/2}}{\sqrt{u}}\right)$&   Multi-class; Corollary \ref{thm:cor} \\
\noalign{\smallskip}\hline
\end{tabular}
\label{tab:ConvRates}
\end{center}
\end{table}

Our two step multiclass SSL approach is in between generative and discriminant approaches, and hence bears similarity with the study of \cite{UrnerShalev-ShwartzBen-David2011}. The main difference is however that the proposed approach does not rely on any pseudo-labeling mechanism and that  our analyzes are based on the Rademacher complexity leading to dimension free data-dependent bounds. On another level and under the PAC-Bayes setting, \cite{Kaeaeriaeinen2005} showed that in the realizable case where the hypothesis set contains  the Bayes classifier,  the obtained excess risk bound takes  the form $\inf\limits_{f\in F_0}\sup\limits_{g\in F_0} \hat d(f,g)
+ \tilde O\left({u^{-1/2}}\right)$; where $\hat d(f,g)$ is a normalized empirical disagreements between two hypothesis that correctly
classify the labeled set and can be of order at least $\tilde O\left({n^{-1/2}}\right)$. The convergence rates of the mentioned bounds are sum up in Table \ref{tab:ConvRates}. From these results, it becomes apparent that the convergence rate deduced from corollary \ref{thm:cor}, (Equation \ref{eq:convrate}) extends those found in \cite{Kaeaeriaeinen2005,Leskes2005} for multiclass classification.

\section{Experimental Results}
\label{sec:exps}

We perform experiments on six publicly available datasets. The three first ones are \texttt{Fungus}, \texttt{Birds} and \texttt{Athletics} that consist of three aggregations of lead nodes that go down from parent nodes in the ImageNet hierarchy\footnote{\url{http://www.image-net.org/challenges/LSVRC/2010/}}. Each image is characterized by a Fisher vector representation as described in \cite{HarchaouiDouzePaulinEtAl2012}. The three others collections are respectively the \texttt{MNIST} database of handwritten digits, the pre-processed 20 Newsgroups (\texttt{20-NG}) collection\footnote{\url{http://www.csie.ntu.edu.tw/~cjlin/libsvmtools/datasets/multiclass.html}} and the \texttt{USPS} dataset\footnote{\url{http://www-i6.informatik.rwth-aachen.de/~keysers/usps.html}}. Table \ref{tab:Char} resumes the characteristics of these datasets. The proportions of training and test sets were kept fixed to those given in the released data files. Within the training set ($\lda\cup\ulda$) we randomly sampled labeled examples $\lda$, with different sizes, and used the remaining as unlabeled data.

\begin{table}[!ht]
\renewcommand\thetable{2}
\caption{Characteristics of datasets used in our experiments.}
\centering
\begin{tabular}{ccccc}
  \hline\noalign{\smallskip}
dataset & $|\lda\cup\ulda|$ & size of the test & dimension, $d$ & \# of classes, $K$ \\
\noalign{\smallskip}\hline\noalign{\smallskip}
\texttt{Birds} & 5785 & 5596 & 4096 & 196 \\
\texttt{Athletics} & 28752 & 28727 & 4096 & 51 \\
\texttt{Fungus} & 50270 & 50271 & 4096 & 134 \\
\texttt{20-NG} & 15936 & 3393 & 62061 & 20\\ 
\texttt{MNIST} & 60000 & 10000 & 780 & 10 \\
\texttt{USPS} & 7291 & 2007 & 256 & 10 \\ 
\noalign{\smallskip}\hline
\end{tabular}

\label{tab:Char}
\end{table}

To validate the proposed penalized based multiclass semi-supervised learning approach (\SMPL), we compared its results with respect to a multiclass extension of a popular SSL algorithm proposed within each of the Generative, Graphical and Discriminant approaches. More precisely we considered the extension of the label propagation algorithm to the multiclass case (\MLP)  proposed by \cite{WangTuTsotsos2013}. A generative SSL model based on the mixture of gaussians (\SSGM), the extension of TSVM\footnote{\url{http://svmlight.joachims.org/}} \cite{vapnik2000nature} to the multiclass case (\MTSVM), and a purely supervised
technique which does not make use of any unlabeled examples in the
 training stage (\SUP).

As the clustering algorithm $\A$, we employed the Nearest Neighbor Clustering technique proposed in \cite{BubeckLuxburg2009}, and fixed $m=4K$, $\kappa=2$ and $\eta=10^{-3}$. Meaning that each cluster in $\C_\kappa(\eta)$ is mainly composed of the two most predominant classes within it. For the second stage of {\SMPL}, as well as for {\SUP} and {\MTSVM}, we adapted the aggregated one-versus-all approach using a linear kernel {\SVM} that respects the conditions of corollary \ref{thm:cor}. The penalized objective function can be easily implemented using convex optimization tools for convex surrogates of the 0/1 loss. The parameter $C$ of the {\SVM} classifier is determined by five fold cross-validation  in logarithmic range between $10^{-4}$ and $10^4$ over the available labeled training data. Results are evaluated over the test set using the accuracy,  and the reported performance
is averaged over $25$ random (labeled/unlabeled/test) sets of the initial collections.

\begin{table}[b!]
\renewcommand\thetable{3}
\caption{Means and standard deviations of the classification accuracy on test data over the $25$ trials for each data set. $n_y$ refers to the average number of labeled examples per class in each data set.  $^\fleche$ indicates statistically significantly worse performance than the best result,  shown in bold, according to a Wilcoxon rank sum test ($p < 0.05$) \cite{Lehmann}.}
\begin{center}
{\small
\begin{tabular}{l c c |  c c c  c  c}
\hline\noalign{\smallskip}
Dataset            & $n_y$   &   $n/(n+u)$  &  \SUP & \SMPL & \MLP     &   \SSGM&   \MTSVM    \\
\noalign{\smallskip}\hline\noalign{\smallskip}
\texttt{Birds}         & $5$   & $0.18$ & $.294^\fleche${\tiny $\pm.03$} & $\mathbf{.344}${\tiny $\pm.03$} & $.303^\fleche${\tiny $\pm.06$} & $.286^\fleche${\tiny $\pm.08$} & $.312^\fleche${\tiny $\pm.04$}\\
\texttt{Athletics}     & $43$  & $0.08$ & $.258^\fleche${\tiny $\pm.03$} & $\mathbf{.273}${\tiny $\pm.02$} & $.259^\fleche${\tiny $\pm.05$} & $.246^\fleche${\tiny $\pm.07$} & $.263${\tiny $\pm.04$}\\
\texttt{Fungus}        & $15$  & $0.04$ & $.121^\fleche${\tiny $\pm.03$} & $\mathbf{.160}${\tiny $\pm.03$} & $.125^\fleche${\tiny $\pm.06$} & $.107^\fleche${\tiny $\pm.05$} & $.134^\fleche${\tiny $\pm.04$}\\
\texttt{20-NG} & $16$  & $0.02$ & $.468^\fleche${\tiny $\pm.05$} & $\mathbf{.531}${\tiny $\pm.03$} & $.476^\fleche${\tiny $\pm.06$} & $.452^\fleche${\tiny $\pm.04$} & $.484^\fleche${\tiny $\pm.04$}\\
\texttt{MNIST}         & $120$ & $0.02$ & $.767^\fleche${\tiny $\pm.03$} & $\mathbf{.799}${\tiny $\pm.02$} & $.771^\fleche${\tiny $\pm.05$} & $.758^\fleche${\tiny $\pm.06$} & $.781^\fleche${\tiny $\pm.01$}\\
\texttt{USPS}          & $14$  & $0.02$ & $.790^\fleche${\tiny $\pm.03$} & $\mathbf{.821}${\tiny $\pm.02$} & $.796^\fleche${\tiny $\pm.04$} & $.788^\fleche${\tiny $\pm.06$} & $.801^\fleche${\tiny $\pm.02$}\\
\noalign{\smallskip}\hline
\end{tabular}
}
\end{center}
\label{tab:Accuracy}
\end{table}

Table \ref{tab:Accuracy} summarizes results obtained by \SUP, \SMPL, \MLP, {\SSGM} and {\MTSVM} when a very small proportion of labeled training data is used in the learning of the models.  We use boldface to indicate the highest performance rates, and the symbol $^\fleche$ indicates that performance is significantly worse than the best result, according to a Wilcoxon rank sum test used at a p-value threshold of 0.05 \cite{Lehmann}.  From these results it becomes clear that 
\begin{itemize}
\item[-] The algorithm {\SMPL} performs significantly better than all of the  four other algorithms, and it improves over {\SUP} by an average of 1.5 to 6.5\% on different datasets.
\item[-] {\MLP} and {\MTSVM} also perform better than {\SUP}, though not in the same range than previously, while the mixture of Gaussians {\SSGM} does worse than {\SUP} especially in the cases where the dimension of the problem is high. 
\item[-] Finally, the difference in performance between {\SMPL} and {\MTSVM} is smaller than the one between the former and {\MLP}.
\end{itemize}

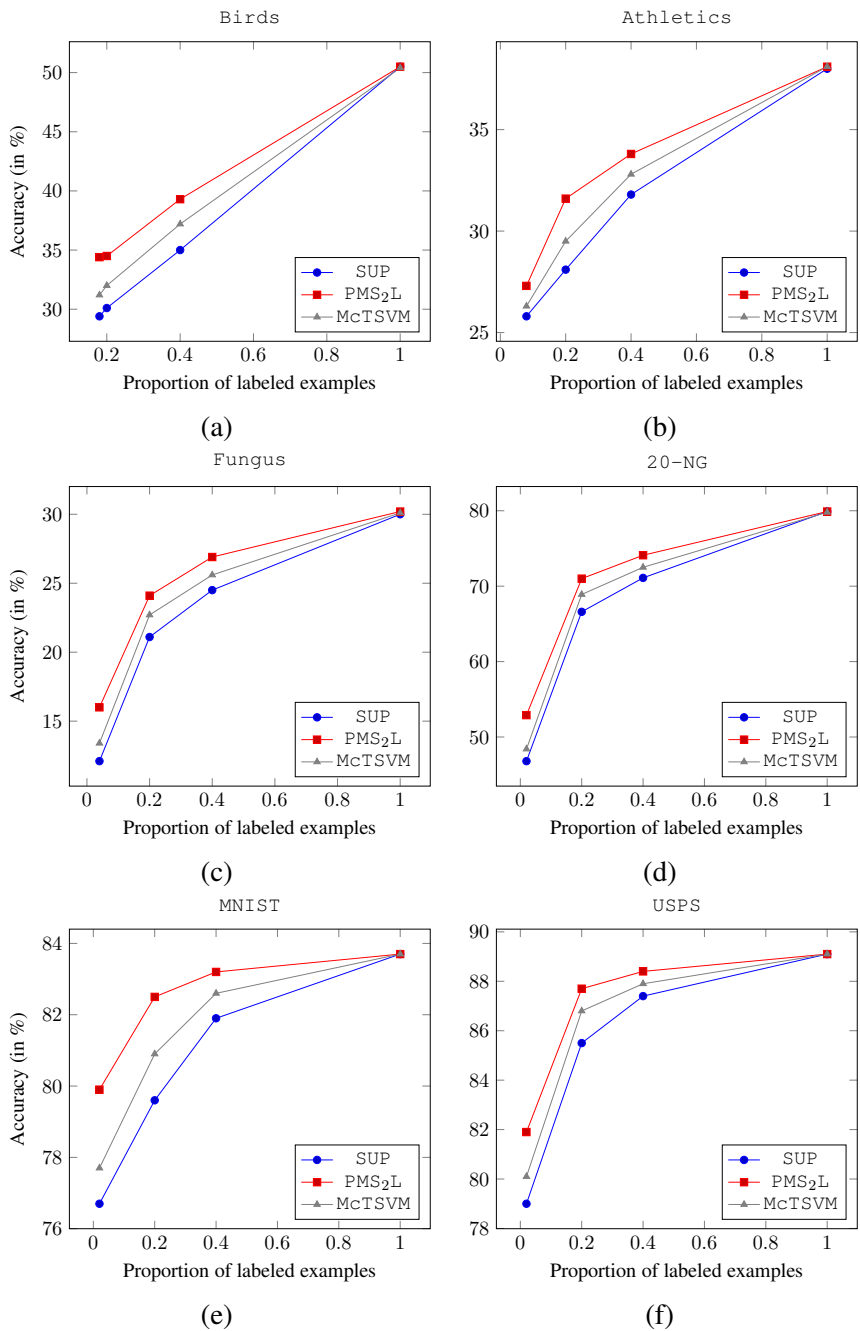
\begin{figure}[!htb]
  \centering
\renewcommand\thetable{1}
\caption{Accuracy in percentage with respect to the proportion of labeled examples in the initial 
training set for ImageNet \texttt{Birds} (a), \texttt{Athletics} (b), \texttt{Fungus} (c);  \texttt{20-NG} (d), \texttt{MNIST} (e), and \texttt{USPS} (f). Each reported performance on the test is averaged over $25$ random (labeled/unlabeled/test) sets of the initial collections. }
\begin{tabular}{cc}
        \begin{tikzpicture}[scale=0.7]
    \begin{axis}[legend pos=south east, title=\texttt{Birds},xlabel=Proportion of labeled examples,ylabel=Accuracy (in \%)]
         \addplot table [x=a, y=b, col sep=comma] {exp-birds.txt};
	  \addlegendentry{\SUP}
         \addplot table [x=a, y=c, col sep=comma] {exp-birds.txt};
         \addlegendentry{\SMPL}
         \addplot[color=gray, mark=triangle*] table [x=a, y=d, col sep=comma] {exp-birds.txt};
       \addlegendentry{\MTSVM}
    \end{axis}
    \end{tikzpicture}  
    &
    \begin{tikzpicture}[scale=0.7]
    \begin{axis}[legend pos=south east,title=\texttt{Athletics},xlabel=Proportion of labeled examples]
         \addplot table [x=a, y=b, col sep=comma] {exp-athletics.txt};
	  \addlegendentry{\SUP}
         \addplot table [x=a, y=c, col sep=comma] {exp-athletics.txt};
         \addlegendentry{\SMPL}
        \addplot[color=gray, mark=triangle*] table [x=a, y=d, col sep=comma] {exp-athletics.txt};
       \addlegendentry{\MTSVM}
    \end{axis}
    \end{tikzpicture} 
    \\
    (a)&(b) \\
  \begin{tikzpicture}[scale=0.7]
    \begin{axis}[legend pos=south east, title=\texttt{Fungus},xlabel=Proportion of labeled examples,ylabel=Accuracy (in \%)]
         \addplot table [x=a, y=b, col sep=comma] {exp-fungus.txt};
	  \addlegendentry{\SUP}
         \addplot table [x=a, y=c, col sep=comma] {exp-fungus.txt};
         \addlegendentry{\SMPL}
         \addplot[color=gray, mark=triangle*] table [x=a, y=d, col sep=comma] {exp-fungus.txt};
       \addlegendentry{\MTSVM}
    \end{axis}
    \end{tikzpicture}
&
       \begin{tikzpicture}[scale=0.7]
        \begin{axis}[legend pos=south east,title=\texttt{20-NG},xlabel=Proportion of labeled examples]
         \addplot table [x=a, y=b, col sep=comma] {exp-news20.txt};
	  \addlegendentry{\SUP}
         \addplot table [x=a, y=c, col sep=comma] {exp-news20.txt};
         \addlegendentry{\SMPL}
           \addplot[color=gray, mark=triangle*] table [x=a, y=d, col sep=comma] {exp-news20.txt};
       \addlegendentry{\MTSVM}
    \end{axis}
    \end{tikzpicture}
    \\
(c)&(d)\\
    \begin{tikzpicture}[scale=0.7]
      \begin{axis}[legend pos=south east, title=\texttt{MNIST},xlabel=Proportion of labeled examples,ylabel=Accuracy (in \%)]
         \addplot table [x=a, y=b, col sep=comma] {exp-mnist.txt};
	  \addlegendentry{\SUP}
         \addplot table [x=a, y=c, col sep=comma] {exp-mnist.txt};
         \addlegendentry{\SMPL}
           \addplot[color=gray, mark=triangle*] table [x=a, y=d, col sep=comma] {exp-mnist.txt};
       \addlegendentry{\MTSVM}
    \end{axis}
    \end{tikzpicture}
 &
    \begin{tikzpicture}[scale=0.7] 
    \begin{axis}[legend pos=south east,title=\texttt{USPS},xlabel=Proportion of labeled examples]
         \addplot table [x=a, y=b, col sep=comma] {exp-usps.txt};
	  \addlegendentry{\SUP}
         \addplot table [x=a, y=c, col sep=comma] {exp-usps.txt};
         \addlegendentry{\SMPL}
           \addplot[color=gray, mark=triangle*] table [x=a, y=d, col sep=comma] {exp-usps.txt};
       \addlegendentry{\MTSVM}
    \end{axis}
    \end{tikzpicture}\\
(e)&(f)
    \end{tabular}
\label{fig:Comparaison}
 \end{figure}

Our analysis of these results is that the Nearest Neighbor Clustering technique \cite{BubeckLuxburg2009} is effectively able to map correctly the considered data, into homogenous clusters containing mostly unlabeled examples of the same class than the $\kappa=2$ most predominant classes contained in them. In this case, the penalized term of the objective function used to learn the classifier (Equation \ref{eq:PEL}) forcefully helps to pick a better hypothesis in the set of linear classifiers, than when only labeled training data are used. Hence, for unlabeled examples within a given cluster, the constraint of predicting the same classes than the $\kappa=2$ most predominant classes of that cluster makes the decision boundary to pass through regions where the unsigned margins of unlabeled examples are small. As stated in section \ref{sec:discussion}, this is exactly how \texttt{TSVM} works, and the proximity of results between {\MTSVM} and {\SMPL}, compared to the two other SSL algorithms can be explained by the similitude of the assumptions leading to the development of these models. 

However, the fundamental difference between these two algorithms in the iterative pseudo-labeling of unlabeled examples (or not), would do that, when the proportion of labeled training data is small, the iterative pseudo-labeling steps of {\MTSVM} injects noise into the learning process at the same level or even more than the true labeled information.  The question therefore arises as to how these two techniques behave for more labeled training data available at the learning phase? 

 In order to analyze more finely this situation,  we compared {\SUP}, {\SMPL} and {\MTSVM} for an increasing size of the labeled training data. Figure \ref{fig:Comparaison}, illustrates this by showing the accuracy (in percentage) with respect to the number of labeled examples  in the initial labeled training set $\lda$. The main observations drawn from these results, are:
 \begin{itemize}
 \item[-] As expected, all performance curves increase monotonously  with respect to the additional labeled data and converge to the same performance. We note that when all the labeled training data are used for learning the linear SVM gives the same results than those reported in the state-of-the art (e.g. the MLP model with no hidden layer on \texttt{USPS} \cite{LeCun2001} and \cite{Maji:EECS-2009-159}).
 \item[-] Though {\MTSVM} takes advantage of unlabeled data in its learning process, it is outperformed by {\SMPL}.  
 \item[-] On ImageNet \texttt{Birds} and \texttt{MNIST}, a non-negligible quantity of labeled examples is necessary for {\SUP} to catch the performance of {\SMPL} learned with the same proportion of labeled data than the one of Table \ref{tab:Accuracy}, and the remaining unlabeled training data.
 \end{itemize}
  These behaviour first suggest that when enough labeled data is available, unlabeled data do not serve the learning algorithm as for the reverse situation. These results suggest that for SSL discriminant techniques designed following the low density separation hypothesis,  a more convenient approach than the pseudo-labeling strategy, used in most of these techniques, would  be the incorporation of a penalized factor concerning unlabeled examples into the objective of the learning algorithm as the one proposed in Equation \ref{eq:PEL}.

%
%
%
%

\section{Conclusion}
\label{sec:concl}
The contributions of this paper are twofold. First, we proposed a bound on the risk of a  multiclass classifier trained over partially labeled training data. We derived data-dependent bounds for the generalization error of a classifier trained by minimizing an objective function that consists of an empirical risk term, estimated over the labeled training set, and a penalized term corresponding to the ratio of unlabeled examples of each cluster; within the $\kappa$ bounded set of clusters, for which their predicted class does not belong to the set of the associated $\kappa$ predominant classes. The analysis of this bound for kernel-based hypotheses reveals a convergence rate that is an extension to the multiclass case, of some other rates over the bounds of the excess risk proposed in the  literature. Empirical results on a various datasets support our findings by showing that the proposed algorithm is competitive compared to different extensions of binary semi-supervised learning algorithms and that it  may significantly increase classification performance in the most interesting situation, when there are few labeled data available for training.

\section*{Acknowledgments}
The authors are thankful to the anonymous reviewers for their numerous helpful suggestions which significantly improved the paper. This work has been partially supported by the THANATOS project funded by \textit{Appel \`a projets Grenoble Innovation Recherche}.  
The work of YM at LANL was funded  by DOE/GMLC  2.0  project:  ``Emergency  Monitoring  and  controls  through  new technologies  and  analytics''.

%
%

\appendix
\section{Mathematical Tools}
\label{sec:Appendix}

\begin{thrm}[McDiarmid's inequality]
\label{thm:McDiarmid}
Let $X_1, \ldots, X_u\in \X^u$ be a set of $u \ge 1$ independent random variables and assume that there exist $c_1,\ldots, c_u > 0$ such that $\phi: \X^u\rightarrow \RR$ satisfies the following condition:
\[
|\phi(x_1, \ldots,  x_i, \ldots, x_u) - \phi(x_1, \ldots, x^{'}_i, \ldots, x_u)|\leq c_i , 
\]
for all $i \in [\![1,u]\!]$ and any points $x_1, \ldots , x_u, x^{'}_i\in \X$. Let $\phi(S)$ denote $\phi(X_1, \ldots, X_u)$,
then, for all $\epsilon > 0$, the following inequalities hold:
\begin{align*}
\PP[\phi(S) -\EE[\phi(S)] \geq \epsilon] &\leq \exp\left(\frac{-2\epsilon^2}{\sum_{i=1}^u c^2_i}\right),\\ 
\PP[\phi(S) - \EE[\phi(S)] \leq −\epsilon] & \leq \exp\left(\frac{-2\epsilon^2}{\sum_{i=1}^u c^2_i}\right).
\end{align*}
\end{thrm}

\begin{thrm}[Minimal matching distance]
\label{thm:DeltaDist}

Let $\A_{Z_1}$ and $\A_{Z_2}$ be two partitions obtained by a clustering algorithm  $\A$ over two finite sets $Z_1$ and $Z_2$. Then for any sample set  $\tilde Z\subseteq \X$, of size $n$, where $\forall \obs\in\tilde Z, \A_{Z}(\obs)\in\{1,\ldots,G\}$ is the partition of $\obs$;  the function
\begin{equation*}
\Delta_n: (\A_{Z_1}, \A_{Z_2},\tilde Z) \mapsto \mathop{\min}_{\pi} \frac{1}{n}\sum\limits_{\obs\in \tilde Z} \Indicator{\A_{Z_1}(\obs)\neq \pi(\A_{Z_2}(\obs))},
\end{equation*}

is a metric over the space of clusterings. 
\end{thrm}
\begin{Proof}
For all $\A_{Z_1}, \A_{Z_2}, \A_{Z_3}$, and $\tilde Z$ the following conditions are indeed satisfied :
\begin{enumerate}
\item \textit{non-negativity: } $\Delta_n(\A_{Z_1}, \A_{Z_2},\tilde Z)\geq 0$,
\item \textit{identity: } $\Delta_n(\A_{Z_1}, \A_{Z_2},\tilde Z)= 0 \Leftrightarrow \A_{Z_1}= \A_{Z_2}$,
\item \textit{symmetry: } $\Delta_n(\A_{Z_1}, \A_{Z_2},\tilde Z)= \Delta_n(\A_{Z_2}, \A_{Z_1},\tilde Z)$,
\item \textit{triangle inequality: } $\Delta_n(\A_{Z_1}, \A_{Z_2},\tilde Z)\leq \Delta_n(\A_{Z_1}, \A_{Z_3},\tilde Z) + \Delta_n(\A_{Z_2}, \A_{Z_3},\tilde Z)$.
\end{enumerate}
The last inequality is due to the fact that for any permutations $\pi, \pi_1$ and $\pi_2$, we have ~:
\[
\forall \obs\in\tilde Z, \Indicator{\A_{Z_1}(\obs)\neq \pi(\A_{Z_2}(\obs))} \le \Indicator{\A_{Z_1}(\obs)\neq \pi_1(\A_{Z_3}(\obs))} + \Indicator{\A_{Z_3}(\obs)\neq \pi_2(\A_{Z_2}(\obs))},
\]
summing over all $\obs \in \tilde Z$ gives:
\[
\frac{1}{n}\sum\limits_{\obs \in \tilde Z}\Indicator{\A_{Z_1}(\obs)\neq \pi(\A_{Z_2}(\obs))} \le \frac{1}{n}\sum\limits_{\obs \in \tilde Z} \Indicator{\A_{Z_1}(\obs)\neq \pi_1(\A_{Z_3}(\obs))} + \frac{1}{n}\sum\limits_{\obs \in \tilde Z} \Indicator{\A_{Z_3}(\obs)\neq \pi_2(\A_{Z_2}(\obs))}.
\]

As the last inequality is valid for any permutations $\pi, \pi_1$ and $\pi_2$ over $\tilde Z$ we have~:
\begin{align*}
     \Delta_n(\A_{Z_1}, \A_{Z_2},\tilde Z) &= \min\limits_{\pi} \frac{1}{n} \sum\limits_{\obs \in \tilde Z}\Indicator{\A_{Z_1}(\obs)\neq \pi(\A_{Z_2}(\obs))}  \\ 
     & \le \min\limits_{\pi_1}\frac{1}{n}\sum\limits_{\obs \in \tilde Z} \Indicator{\A_{Z_1}(\obs)\neq \pi_1(\A_{Z_3}(\obs))} + \min\limits_{\pi_2}\frac{1}{n}\sum\limits_{\obs \in \tilde Z} \Indicator{\A_{Z_3}(\obs)\neq \pi_2(\A_{Z_2}(\obs))} \\ 
     & = \Delta_n(\A_{Z_1}, \A_{Z_3},\tilde Z) + \Delta_n(\A_{Z_2}, \A_{Z_3},\tilde Z).
\end{align*}

\end{Proof}

\begin{thrm}[Data-dependent Bennett's inequality, Th. 4, \cite{MaurerPontil2009}]
\label{thrm:maurer}
Let $X$, $X_1$, $\dots$, $X_n$ be i.i.d. random variables with values in $[0, 1]$ and let $\delta > 0$. Then with probability at least $1-\delta$ in $(X_1,\dots,X_n)$ we have
\[\EE\;[X] - \frac{1}{n}\sum\limits_{i=1}^n X_i \le \sqrt{\frac{2 {V}_n(X)\log \frac{2}{\delta}}{n}} + \frac{7\log \frac{2}{\delta}}{3(n-1)},\]
where $V_n(X)$ is the sample variance 
\[V_n(X) = \frac{1}{n(n-1)}\sum_{1\le i < j \le n}(X_i - X_j)^2\]
\end{thrm}

\begin{lemma}[Lemma 8.1, \cite{MohriRostamizadehTalwalkar2012}]
\label{lem:max-rad}
Let $\F_1, \dots, F_l$ be $l$ hypothesis sets in $\RR^\Inp$, $l\geq 1$, and let ${\cal G} = \{\max(h_1, \dots, h_l): h_i \in \F_i\}$, $1\leq i \leq l$. 
Then, for any sample $S$ of size $n$, the empirical Rademacher complexity of $G$ can be upper bounded as follows:
\[\fR_n^*({\cal G}) \le \sum_{i=1}^l \fR_n^*(\F_i).\]
\end{lemma}

%
%
%

\begin{thrm}[Rademacher generalization bounds, Th. 8.1 \cite{MohriRostamizadehTalwalkar2012}]
\label{thrm:rademacher}
Let $G$ be a family of functions mapping from $\Inp$ to $[0,1]$. Then for any $1 > \delta > 0$, with probability at least $1-\delta$ we have for all $g\in G$~:
\[
    \mathbb{E}[g] \le \frac{1}{n}\sum_{i=1}^n g(z_i) + \fR_n^*(G) + 3\sqrt{\frac{\log \frac{2}{\delta}}{2n}}
\]
\end{thrm}


\begin{ddf}[L-regular loss, definition 2 \cite{lei2015multi}]
A loss function $\ell$ is said to be $L$-regular if~:
\begin{enumerate}
\item $\ell(t)$ bounds the 0-1 loss from above: $\ell(t) \ge 1_{t\le 0}$;
\item $\ell$ is $L$-Lipschitz in the sense $|\ell(t_1) - \ell(t_2)| \le L|t_1 - t_2|$;
\item $\ell(t)$ is decreasing and it has a zero point $c_\ell$, i.e., $\ell(c_\ell) = 0$.
\end{enumerate}
\end{ddf}

\begin{thrm}[Multi-class Rademacher generalization bounds; remark 6 \cite{lei2015multi}]
\label{thrm:lei}
~\\
Let~$\F_{\cal H}~\subset \RR^{\Inp \times \Outp}$ be a hypothesis class with $\Outp = \{1,\dots, K\}$. Let $\ell$ be a $L$-regular loss function and denote $B_\ell \doteq \sup_{(\obs,y),h} \ell(m_h(\obs, y))$. 

Suppose that the examples $\lda = \{(\obs_i, y_i); i\in\{1,\ldots, n\}\}$ are i.i.d with respect to a fixed yet unknown probability distribution defined on $\Inp\times\Outp$.
Then, for any $\delta > 0$, with probability at least $1 - \delta$, the following multi-class classification generalization bound holds for any $h\in H$:
\[
R(h) \le 
    \frac{1}{n}\sum_{i=1}^n \ell(m_h(\obs_i,y_i)) + 
    2LK\fR_n^*(\F_{\cal H}) + 
    3B_\ell \sqrt{\frac{\log \frac{2}{\delta}}{2n}},
\]
where $\F_\HHH = \{f:\obs\mapsto h(\obs,y): y\in\Outp, h\in\HHH\}$.
\end{thrm}
Note that,  up-to a constant similar bounds were obtained in  \cite{kuznetsovrademacher} and \cite{maximov2016tight}.
\section{Full proofs}
\label{sec:Proofs}
\begin{customlemma}{1}
Let $\lda=\left(\obs_i,y_i\right)_{1\leq i \leq n}$ and $\ulda=(\obs_{n+i})_{1\leq i \leq u}$ be a labeled and an unlabeled training sets drawn i.i.d. according respectively to a probability distribution $\D$ over $\Inp \times \Outp$,  and its marginal $\D_\Inp$. For any $1>\delta>0$ and any stable clustering algorithm $\A$ that obeys the bounded differences property with constant $L>0$, the following inequality holds  with probability at least $1-\delta$~: 
\begin{equation*}
\Delta_n(\A_{\ulda}, \A^\star, \lda) \le \frac{L}{u} + L\sqrt{\frac{\log \frac{2}{\delta}}{2u}} + \sqrt{\frac{ \log \frac{2}{\delta}}{2n}}.
\end{equation*}

\end{customlemma}

\bigskip

%
%
\noindent \textbf{Proof.} As the function $\Delta_n$ (Eq. \ref{eq:Distance}) is a metric (Appendix, Th. \ref{thm:DeltaDist}); for any labeled training set $\lda\subseteq (\X\times\Y)^n$ and any cluterings $\A_{Z}, \A_{Z'}$ found by the algorithm $\A$ over the sets $Z, Z'$, we have by the triangle inequality~:
\begin{gather*}
    \Delta_n(\A_{Z}, \A^\star, \lda) \le \Delta_n(\A_{Z}, \A_{Z'}, \lda) + \Delta_n(\A_{Z'}, \A^\star, \lda),
\end{gather*}
 hence by the non-negativity of the distance function we have~:
\begin{gather}
\label{eq:subadd}
   \left| \Delta_n(\A_{Z}, \A^\star, \lda) - \Delta_n(\A_{Z'}, \A^\star, \lda)\right| \le \Delta_n(\A_{Z}, \A_{Z'}, \lda).
\end{gather}

Consider the following multivariate function defined over unlabeled training sets of size $u$;
\begin{align*}
\phi: \X^u &\rightarrow \RR \\
Z&\mapsto  \EE_{\lda\sim \D^n}[\Delta_n(\A_Z,\A^\star,\lda)].
\end{align*}
For any unlabeled training sets, $S_u$ and $S'_u$ drawn i.i.d. with respect to the marginal $\D_\X$ that differ only in one observation we have~:
\begin{align}
    \left|\phi(S_u) - \phi(S'_u)\right| &= \left|\EE_{\lda\sim \D^n}\left(\Delta_n(\A_{S_u},\A^\star,\lda) - \Delta_n(\A_{S'_u},\A^\star,\lda)\right)\right| \nonumber \\ 
    &\le  \EE_{\lda\sim \D^n} \left|\left(\Delta_n(\A_{S_u},\A^\star,\lda) - \Delta_n(\A_{S'_u},\A^\star,\lda)\right)\right| \label{eq:one}\\ 
    &\le  \EE_{\lda\sim \D^n}\!\! \left[\Delta_n(\A_{S_u}, \A_{S'_u}, \lda)\right]\!\! =\Delta(\A_{S_u}, \A_{S'_u})\le \frac{L}{u}. \label{eq:three}
\end{align}
where (Eq. \ref{eq:one}) is due to the triangle inequality with absolute value; and  (Eq. \ref{eq:three}) results from (Eq. \ref{eq:subadd}) and the bounded-difference property of algorithm $\A$ (Eq. \ref{eq:bounded-diff-prop}).

Then by McDiarmid's inequality (Appendix, Th. \ref{thm:McDiarmid}) for any $\epsilon > 0$ we get~:

\begin{gather*}
    \PP\left[\phi(\ulda) - \EE_{\ulda\sim \D_\X^u} \phi(\ulda) \ge \epsilon\right] \le \exp\left(-\frac{2\epsilon^2 u}{L^2}\right)
\end{gather*}
Setting the right-hand side to be $\delta/2$, and solving for $\epsilon$, we obtain that with probability at least $1-\frac{\delta}{2}$~:
\begin{equation}
\phi(\ulda) 
\le \EE_{\ulda\sim \D_\X^u} [\phi(\ulda)] + L \sqrt{\frac{\log \frac{2}{\delta}}{2u}} 
\le \frac{L}{u} + L \sqrt{\frac{\log \frac{2}{\delta}}{2u}}. \label{eq:BndDiff1}
\end{equation}

Where the last inequality is due to the stability of the clustering algorithm $\A$ (Eq. \ref{eq:consistency-prop}).
Furthermore, by bounding $\phi(\ulda) = \EE_{\lda\sim \D^n}[\Delta_n(\A_\ulda,\A^\star,\lda)]$ in terms of $\lda$ using again the McDiarmid  inequality we have for any $\epsilon > 0$~:
\begin{gather*}
    \PP\left[\Delta_n(\A_{\ulda}, \A^\star, \lda) - \phi(\ulda) \ge \epsilon\right] \le e^{-2n\epsilon^2},
\end{gather*}
Indeed, if we consider the multivariate function $\psi: \lda\mapsto \Delta_n(\A_{\ulda}, \A^\star, \lda)$; changing a single labeled observation in $\lda$ could not change $\Delta_n(\A_{\ulda}, \A^\star, \lda)$ on more than $1/n$ by definition (Eq. \ref{eq:Distance}). Hence, by setting the right-hand side to be $\delta/2$, and solving for $\epsilon$, we obtain that with probability greater than $1-\frac{\delta}{2}$~:
\begin{gather}\label{eq:bound2}
 \Delta_n(\A_{\ulda}, \A^\star, \lda) \le \phi(\ulda) + \sqrt{\frac{\log \frac{2}{\delta}}{2n}}.
\end{gather}

Applying the union bound on both inequalities (Eq. \ref{eq:BndDiff1}) and (Eq. \ref{eq:bound2}), we finally get that for any labeled and unlabeled training sets $\lda$ and $\ulda$ and with probability at least $1-\delta$~:
\[
\hspace{4cm}\Delta_n(\A_\ulda, \A^\star, \lda) \le \frac{L}{u} + L\sqrt{\frac{\log \frac{2}{\delta}}{2u}} + \sqrt{\frac{\log \frac{2}{\delta}}{2n}}.\hspace{4cm}\square
\]

\begin{customlemma}{2}
Let $\HHH \subseteq \mathbb{R}^{\Inp\times \Outp}$ be a hypothesis set where $\Outp=\{1,\ldots,K\}$, and let $\lda=\left(\obs_i,y_i\right)_{1\le i \le n}$ and $\ulda=(\obs_{n+i})_{1\le i \le u}$ be two sets of labeled and unlabeled training data, drawn i.i.d. respectively according to a probability distribution over $\Inp \times \Outp$ and a marginal distribution $\D_{\X}$. Fix $\rho>0$, $\kappa\in\{1,\ldots,K\}$ 
then for any $1>\delta > 0$, the following multiclass classification generalization error bound holds with probability at least $1-\delta$ 
for all $h\in \HHH$ learned by algorithm \ref{alg:smplpca} over a single $\kappa$-uniformly bounded cluster $\C_j\in \C_\kappa(\eta)$ derived from $\ulda$ by a clustering algorithm $\A_\ulda$ that partitions the input space into $G$ clusters~:
    \begin{align*}
     R(h, \C_j) & \leq {\widehat R}_{\rho}(h, \C_j) + \frac{\eta}{G} + \frac{2\kappa}{\rho}\fR_{n,j}^*(\F_\HHH) + \frac{2K}{\rho} \fR^*_{u,j}(\F_\HHH)  \\ 
     &\quad + 5\sqrt{\frac{\kappa \neta(j)\log \frac{8K}{\delta}}{2n^2}} + 5\sqrt{\frac{\kappa \ueta(j)\log \frac{8K}{\delta}}{2u^2}} + \frac{7\log \frac{8}{\delta}}{3(n-1)}  + \frac{7\log \frac{8}{\delta}}{3(u-1)},
   \end{align*}
\noindent 
where $\neta(j) = |\lda \cap \C_j|$, $\ueta(j) = |\ulda \cap \C_j|$, $\fR_{n,j}^* = \EE_{\sigma} \sup\limits_{f\in\F_\HHH} \frac{2}{n}
\sum_{\obs_i\in\lda\cap\C_j}\sigma_i f(\obs_i)
$, and  $\fR_{u,j}^* = \EE_{\sigma} \sup\limits_{f\in\F_\HHH} \frac{2}{u}
\sum_{\obs_i\in\ulda\cap\C_j}\sigma_i f(\obs_i)
$.
\end{customlemma}

\bigskip

\begin{Proof}
We start with the decomposition of the risk estimated in a single $\kappa$-uniformly bounded cluster $\C_j\in \C_\kappa(\eta)$, by considering two situations where the prediction $\mu_h(\obs) = \arg\max\limits_{y\in \Outp} h(\obs,y)$ falls within any set of confident clusters and without them respectively:
\begin{align}
\label{ineq:l01}
\!\!R(h, \C_j)=\EE [\mu_h(\obs) \neq y \wedge \obs\in \C_j] \le &  \EE_{(\obs, y)\sim\D} [\mu_h(\obs) \neq y \wedge \mu_h(\obs) = \mu_h(\obs, \Outp_\kappa')\wedge \obs\in \C_j] + \nonumber \\
 & \EE_{(\obs, y)\sim\D} [\mu_h(\obs) \neq y \wedge \mu_h(\obs) \neq \mu_h(\obs, \Outp_\kappa')\wedge \obs\in \C_j] 
\end{align}
where $\mu_h(\obs, \Outp_\kappa') = \arg\max\limits_{y\in \Outp_\kappa'} h(\obs,y)$ and $\Outp_\kappa'\subseteq \Outp$, $|\Outp_\kappa'| \le \kappa$. 

The first term in the inequality above involves the margin of examples and it can be upper-bounded using the definition of the $\rho$-margin loss (Eq. \ref{eq:RhoMArgin}) estimated over the labeled examples that are in cluster $\C_j$~:
\begin{align}
\label{ineq:d0}
\!\!\!\!\EE_{(\obs, y)\sim\D} [\mu_h(\obs)\! \neq\! y \wedge \mu_h(\obs) \!= \!\mu_h(\obs, \Outp_\kappa')\!\wedge\! \obs\in \C_j] &\! = \!\EE_{(\obs, y)\sim\D} [\mu_h(\obs, \Y_\kappa') \!\neq \!y\! \wedge\! \obs\in \C_j]  \nonumber \\ & \le
\EE_{\lda\sim\D^n} [\Phi_{\rho}(m_h(\obs, y, \Y_\kappa')) \!\wedge\! \obs\!\in\! \C_j],
\end{align}
where $m_h(\obs, y, \Y_\kappa')= h(\obs,y) -  \max_{y'\in \Y_\kappa'\setminus\{y\}} h(\obs,y'),  \obs\in\C_j$.

 Expected risk over a single cluster $\C_j$ can be decomposed through conditional risk as~:
\begin{align}
\label{ineq:d01a}
\EE_{\lda\sim\D^n} [\Phi_{\rho}(m_h(\obs, y, \Outp'_\kappa)) \wedge \obs\in \C_j] ~=~ & \EE_{\lda\sim\D^n} [\Phi_{\rho}(m_h(\obs, y, \Outp'_\kappa)) \bigl| \obs\in \C_j] \times \nonumber \\
 &   \EE_{\lda\sim\D^n} [\obs \in \C_j]
\end{align}
From the data-dependent Bennett's inequality (appendix A, theorem \ref{thrm:maurer}), we have with probability at least $1-\delta/4$~:
\begin{align}
\label{ineq:d01b}
\EE_{\lda\sim\D^n} [\obs \in \cap\C_j] \le \frac{n_\eta(j)}{n} + \sqrt{\frac{2n_\eta(j)\log \frac{8}{\delta}}{n^2}} + \frac{7 \log \frac{8}{\delta}}{3(n-1)},
\end{align}
where $\neta(j)=|\lda\cap\C_j|$, and the sample variance, which is upper-bounded by~:
\[
V_n(\obs\in \C_j) = \frac{n_\eta(j)(n-n_\eta(j))}{n(n-1)} \le \frac{n_\eta(j)}{n}.
\]

Since $0\le \Phi_{\rho}(\cdot) \le 1$ and so $0\le \EE_{(\obs, y)\sim\D} [\Phi_{\rho}(m_h(\obs, y, \Outp'_\kappa)) \bigl| \obs\in \lda\cap\C_j]\le 1$, we have from \eqref{ineq:d01a} and \eqref{ineq:d01b} with probability at least $1-\delta/4$~:
\begin{align}
\label{ineq:d01c}
\EE_{\lda\sim\D^n} [\Phi_{\rho}(m_h(\obs, y, \Outp'_\kappa) \wedge \obs \in \C_j]  \le &~\frac{n_\eta(j)}{n}\EE_{\lda\sim\D^n} [\Phi_{\rho}(m_h(\obs, y, \Outp'_\kappa)) \bigl| \obs\in \C_j] + \nonumber \\ 
& ~ \sqrt{\frac{2n_\eta(j)\log \frac{8}{\delta}}{n^2}} + \frac{7 \log \frac{8}{\delta}}{3(n-1)}
\end{align}

Further, the $\rho$-margin loss function $\Phi_\rho(\cdot)$ (Eq. \eqref{eq:RhoMArgin}) is $1/\rho$-Lipschitz, from the multi-class classification generalization bound proposed in \cite{lei2015multi} (appendix A, theorem \ref{thrm:lei}); it then comes that for any fixed set $\Outp'_\kappa \subset \Outp$, $|\Outp'_\kappa|\le \kappa$ and any $1>\delta>0$ 
with probability at least $1-\delta/4K^\kappa$ we have for all $h\in\HHH$~:

\begin{align}
\label{ineq:d01d}
\EE_{\lda\sim\D^n} [\Phi_{\rho}&(m_h(\obs, y, \Outp'_\kappa) \bigl| \obs \in \C_j] \nonumber\\
\le & \frac{1}{n_\eta(j)}\!\!\!\sum_{(\obs, y)\in\lda\cap \C_j}\!\!\!\!\! \Phi_\rho(m_h(\obs, y, \Outp'_\kappa)) + \frac{2\kappa}{\rho} \fR_{n_\eta(j)}^*(\F) + 3\sqrt{\frac{\log \frac{8 K^\kappa}{\delta}}{2n_\eta(j)}} \nonumber \\
\le 
 &  \frac{1}{n_\eta(j)}\!\!\!\sum_{(\obs, y)\in\lda\cap \C_j}\!\!\! \Phi_\rho(m_h(\obs, y, \Outp'_\kappa))   + \frac{2\kappa}{\rho} \fR_{n_\eta(j)}^*(\F) + 3\sqrt{\frac{ \kappa \log \frac{8K}{\delta}}{2\neta(j)}}, 
\end{align}
where, $\fR_{n_\eta(j)}^*(\F)=\EE_{\sigma} \sup\limits_{f\in\F_\HHH} \frac{2}{n_\eta(j)}
\sum_{\obs_i\in\lda\cap\C_j}\sigma_i f(\obs_i)$.

Now for any possible set of $\kappa$ predominant classes $\Y_\kappa$ in $\C_j$, and using the union bound and the inequality  $\sum_{i=1}^k \binom {K}{i} \le 2 K^\kappa$, it comes from \eqref{ineq:d01c} and \eqref{ineq:d01d} and the union bound, we have with probability at least $1-\delta/2$~: 
\begin{align}
\label{ineq:d1}
& \EE_{\lda\sim\D^n} [\Phi_{\rho}(m_h(\obs, y, \Y_\kappa)) \wedge \obs\in \C_j] \le \nonumber \\ & \frac{1}{n}\sum_{(\obs, y)\in\lda\cap \C_j} \Phi_\rho(m_h(\obs, y, \Outp_\kappa))   + \frac{2\kappa}{\rho} \fR_{n,j}^*(\F) + 5\sqrt{\frac{ \kappa \neta(j) \log \frac{8K}{\delta}}{2n^2}} + \frac{7\log \frac{8}{\delta}}{3(n-1)},
\end{align}

Where $\fR_{n,j}^* = \EE_{\sigma} \sup\limits_{f\in\F_\HHH} \frac{2}{n}\sum_{\obs_i\in\lda\cap\C_j}\sigma_i f(\obs_i)$.
 By decomposing the sum in the first term of the above inequality, and considering the two cases where the class label $y$ is within or without  $\Outp_\kappa$~:
 \begin{equation*}
\sum_{(\obs, y)\in\lda\cap \C_j} \Phi_\rho(m_h(\obs, y, \Outp_\kappa))\le \!\!\sum_{(\obs, y)\in\lda\cap \C_j\wedge y\in\Outp_\kappa} \!\!\!\!\!\!\!\!\!\Phi_\rho(m_h(\obs, y, \Outp_\kappa))+\!\!\sum_{(\obs, y)\in\lda\cap \C_j\wedge y\notin\Outp_\kappa} \!\!\! \!\!\!\!\!\!\Phi_\rho(m_h(\obs, y, \Outp_\kappa)),
\end{equation*}

  Here we are in the case where $\mu_h(\obs) = \mu_h(\obs, \Outp_\kappa)$ (Eq. \ref{ineq:l01}) so, $\forall (\obs, y)\in\lda\cap \C_j\wedge y\in\Outp_\kappa,$ $\Phi_\rho(m_h(\obs, y, \Outp_\kappa))=\Phi_\rho(m_h(\obs, y))$, and $\forall (\obs, y)\in\lda\cap \C_j\wedge y\notin\Outp_\kappa, \Phi_\rho(m_h(\obs, y, \Outp_\kappa))\le \Indicator{y \not\in \Outp_\kappa \wedge \obs\in \C_j}$. Hence, for any sample $\lda$ and a set of predominant classes $\Y_\kappa$ we have 
\begin{align*}
\frac{1}{n}\sum_{(\obs, y)\in\lda\cap \C_j} \Phi_\rho(m_h(\obs, y, \Outp_\kappa)) \le & \frac{1}{n}\sum_{(\obs, y)\in\lda\cap \C_j} \Phi_\rho(m_h(\obs, y))+ \frac{1}{n}\sum_{(\obs, y)\in\lda} \Indicator{y \not\in \Outp_\kappa \wedge \obs\in \C_j}\\
\le &  \frac{1}{n}\sum_{(\obs, y)\in\lda\cap \C_j} \Phi_\rho(m_h(\obs, y))+ \frac{1}{n}\sum_{(\obs, y)\in\lda} \Indicator{y \not\in \Outp_\kappa \wedge \obs\in \C_j}.
\end{align*}
From definition \eqref{eq:EtaClusters} we have  $\frac{1}{n}\sum_{(\obs, y)\in\lda} \Indicator{y \not\in \Outp_\kappa \wedge \obs\in \C_j}\le \eta/G$, and so
 \begin{align}
 \label{ineq:l02}
\EE_{(\obs, y)\sim\D} & [\mu_h(\obs) \neq y \wedge \mu_h(\obs) = \mu_h(\obs, \Outp_\kappa)\wedge \obs\in \C_j] ~\le~ \nonumber\\ 
& \frac{1}{n}\sum_{(\obs, y)\in\lda\cap \C_j} \Phi_\rho(m_h(\obs, y)) + \frac{\eta}{G}  + \frac{2\kappa}{\rho} \fR_{n,j}^*(\F) + 5\sqrt{\frac{\kappa \neta(j)\log \frac{8K}{\delta}}{2n^2}} + \frac{7\log \frac{8}{\delta}}{3(n-1)}.
 \end{align}

\bigskip

Further, the second term in inequality \eqref{ineq:l01} for any set $\Outp_\kappa \subset \Outp$, $|\Outp_\kappa|\le \kappa$ can be upperbounded using unlabeled data that are in cluster $\C_j$~: 
\noindent
\begin{align*}
\EE [\mu_h(\obs) \neq y \wedge \mu_h(\obs) \neq \mu_h(\obs, \Outp_\kappa)\wedge \obs\in \C_j] & \le 
\EE_{\ulda\sim\D_\Inp^u} [\mu_h(\obs) \neq \mu_h(\obs, \Outp_\kappa)\wedge \obs\in \C_j]  \\ & \le  \EE_{\ulda\sim\D_\X^u}[\Phi_\rho(m'_h(\obs, \Y_\kappa))\wedge \obs\in \C_j],
\end{align*}
where $m'_h(\obs, \Y_\kappa)= \max_{y\in \Y_\kappa(\C_j)} h(\obs,y) -  \max_{y\in \Y\setminus\Y_\kappa(\C_j)} h(\obs,y)$,  
$\obs\in\C_j$. 

\medskip

As the $\rho$-margin loss has its values in $[0,1]$,  from the standard Rademacher complexity bound
(appendix A, theorem \ref{thrm:rademacher}) over i.i.d. sample $\ulda \cap \C_j$, for any $0>\delta>1$ and $\Outp_\kappa \subseteq \Outp$ it comes  that with probability at least $1-\delta/4$~:

\begin{align}
\label{eq:bound4}
    \EE_{\ulda\sim\D_\X^u}[\Phi_\rho(m'_h(\obs, \Y_\kappa))\bigl| \obs\in \C_j] \le &\frac{1}{\ueta(j)} \sum_{\obs\in\C_j\cap\ulda}\Phi_\rho(m'_h(\obs, \Y_\kappa)) +  \nonumber\\
      \sum\limits_{\C_j \in \C_{\kappa}(\eta)} & \EE_{\sigma} \sup\limits_{f\in{\cal G}^{\C_j}_1 \cup {\cal G}^{\C_j}_2} \frac{2}{\ueta(j)}
    \sum_{\obs_i\in\ulda\cap\C_j}\sigma_i f(\obs_i) + {3}\sqrt{\frac{\log \frac{8K^\kappa}{\delta}}{2\ueta(j)}},
\end{align}
where $\mathcal{G}_1^{\C_j} = \{\max_{y\in \Outp_\kappa(\C_j)}h(\obs,y), h\in \F_{\HHH}\}$  and $\mathcal{G}_2^{\C_j} = \{\max_{y\not\in \Outp_\kappa(\C_j)}h(\obs,y), h\in \F_{\HHH}\}$. Due to the monotonicity of supremum, we have for any $\C_j \in \C_\kappa(\eta)$~:
\begin{align*}
\EE_{\sigma} \sup\limits_{f\in{\cal G}^{\C_j}_1 \cup{\cal G}^{\C_j}_2} \frac{2}{\ueta(j)}
\sum_{\obs_i\in\ulda\cap\C_j}\sigma_i f(\obs_i) & \nonumber \\ 
\le \EE_{\sigma} & \sup\limits_{f\in{\cal G}^{\C_j}_1} \frac{2}{\ueta(j)}
\sum_{\obs_i\in\ulda\cap\C_j}\sigma_i f(\obs_i) + \EE_{\sigma} \sup\limits_{f\in{\cal G}^{\C_j}_2} \frac{2}{\ueta(j)}
\sum_{\obs_i\in\ulda\cap\C_j}\sigma_i f(\obs_i)
\end{align*}

By Lemma \ref{lem:max-rad} (Appendix A)  we have~:
\begin{align*}
\EE_{\sigma} \sup\limits_{f\in{\cal G}^{\C_j}_1} \frac{2}{\ueta(j)}
\sum_{\obs_i\in\ulda\cap\C_j}\sigma_i f(\obs_i) + \EE_{\sigma} \sup\limits_{f\in{\cal G}^{\C_j}_2} \frac{2}{\ueta(j)}
\sum_{\obs_i\in\ulda\cap\C_j}\sigma_i &  f(\obs_i)  \nonumber \\ \le K & \EE_{\sigma} \sup\limits_{f\in \F} \frac{2}{\ueta(j)}
\sum_{\obs_i\in\ulda\cap\C_j}\sigma_i f(\obs_i)
\end{align*}

Hence,
\begin{equation}
\label{ineq:l03}
\EE_{\ulda\sim\D_\X^u}[\Phi_\rho(m'_h(\obs, \Y_\kappa))\bigl| \obs\in \C_j]  \le \frac{1}{\ueta(j)} \sum_{\obs\in\C_j}\Phi_\rho(m'_h(\obs, \Y_\kappa)) + \frac{2K}{\rho} \fR^*_{\ueta(j)}(\F) + 3\sqrt{\frac{\kappa \log \frac{8K}{\delta}}{2\ueta(j)}},
\end{equation}
where $\fR^*_{\ueta(j)}(\F)=\EE_{\sigma} \sup\limits_{f\in \F} \frac{2}{\ueta(j)}\sum_{\obs_i\in\ulda\cap\C_j}\sigma_i f(\obs_i)$. Similarly to \eqref{ineq:d01c} we have with probability at least $1-\delta/4$~:
\begin{align}
\label{ineq:d01e}
\EE_{\ulda\sim\D_{\cal X}^u} [\Phi_{\rho}(m_h'(\obs, y, \Outp'_\kappa) \wedge \obs \in \C_j] \le~ & \frac{u_\eta(j)}{u}\EE_{\ulda\sim\D_{\cal X}^u} [\Phi_{\rho}(m_h'(\obs, y, \Outp'_\kappa)) \bigl| \obs\in \C_j] + \noindent \nonumber \\ 
&  \sqrt{\frac{2u_\eta(j)\log \frac{8}{\delta}}{u^2}} + \frac{7 \log \frac{8}{\delta}}{3(u-1)}
\end{align} 

Thus, by \eqref{ineq:l03} and \eqref{ineq:d01e}, and the union bound we have with probability at least $1-\delta/2$:
\begin{align}
\label{ineq:l04}
\EE_{\obs\sim\D_\X}[\Phi_\rho(m'_h(\obs, \Y_\kappa))  \wedge \obs\in \C_j]  \le~& \frac{1}{u}  \sum_{\obs\in\C_j}\Phi_\rho(m'_h(\obs, \Y_\kappa)) + \frac{2K}{\rho} \fR^*_{u,j}(\F) + \nonumber \\ 
 & {5}\sqrt{\frac{\kappa\ueta(j)\log \frac{8K}{\delta}}{2u^2}} + \frac{7 \log \frac{8}{\delta}}{3(u-1)}
\end{align}

\noindent The statement of the Lemma follows from the inequalities \eqref{ineq:l01}, \eqref{ineq:l02},  \eqref{ineq:l04}, and the union bound.  
\end{Proof}

\begin{customthm}{3}
Let $\HHH \subseteq \mathbb{R}^{\Inp\times \Outp}$ be a hypothesis set where $\Outp=\{1,\ldots,K\}$, and let $\lda=\left((\obs_i,y_i)\right)_{i=1}^n$ and $\ulda=(\obs_i)_ {i=n+1}^{n+u}$ be two sets of labeled and unlabeled training data, drawn i.i.d. respectively according to a probability distribution over $\Inp \times \Outp$ and a marginal distribution $\D_{\X}$. Fix $\rho>0$ and $\kappa\in\{1,\ldots,K\}$, and consider a clustering algorithm $\A$ that  obeys the bounded difference property with constant $L$ and is stable. If the $\kappa$-uniformly bounded clusters found in $\Pi_\ulda$ are such that the confident level $\eta$ satisfies $\eta\leq \Delta_n(\A_\ulda,\A^\star,\lda)$,
then for any $1>\delta > 0$ and all $h\in \HHH$ found by the {\SMPL} algorithm using $\A_\ulda$, the following multiclass classification generalization error bound holds with probability at least $1-\delta$~:
   \begin{equation*}
     R(h) \hspace{-1mm}\leq \hspace{-1mm} {\widehat R}_{\rho}(h)   + \frac{L}{u} +\frac{2K}{\rho} (\fR^*_u(\F_\HHH)+ \fR_n(\F_\HHH)) + \frac{2\kappa}{\rho}\fR^*_n(\F_\HHH)+ \frac{7G\log\hspace{-.3mm}\frac{14G}{\delta}}{3\se} + \sqrt{\frac{\hspace{-.3mm}\log\hspace{-.3mm} \frac{14}{\delta}}{\te}}+9\sqrt{\hspace{-.3mm}\frac{\log \hspace{-.3mm}\frac{14KG}{\delta}}{v_*}},
   \end{equation*}
\noindent 
where 
$\frac{1}{\se}\doteq \left(\frac{2}{n-1}+\frac{1}{u-1}\right), \frac{1}{\te} \doteq \frac{L^2}{u}+\frac{1}{n}, \frac{1}{v_*} \doteq \frac{G\kappa \ueta}{2u^2} + \frac{G \kappa\neta+K(n-\neta)}{2n^2}, \neta = |\lda \cap \C_\kappa(\eta)|$ and $\ueta = |\ulda \cap \C_\kappa(\eta)|$. 
\end{customthm}

\bigskip

\begin{Proof}
Let $\Pi_\ulda = \{\C_1,\dots,\C_G\}$ be a set of disjoint clusters found by $\A_\ulda$. We decompose the risk of a classifier by considering the two exclusive cases whether the misclassification error occurs inside or outside the set of $\eta$-confident clusters~:  
\begin{align}
\label{ineq:x01}
R(h)=\EE_{(\obs,y)\sim\D} [\mu_h(\obs) \neq y]  =& \sum\limits_{\C_j\in \C_\kappa(\eta)} \EE_{(\obs,y)\sim\D} [\mu_h(\obs) \neq y \wedge \obs \in \C_j]~ + \nonumber \\ & \sum\limits_{\C_j\not\in \C_\kappa(\eta)} \EE_{(\obs,y)\sim\D} [\mu_h(\obs) \neq y \wedge \obs \in \C_j].
\end{align}

First, we bound the risk over the set of confident clusters. For any cluster $\C_j$ in $\C_\kappa(\eta)$ and any set of confident clusters $\Outp_\kappa(\C_j)$ within it,  from lemma \ref{lem:rademacher} we have with probability at least $1-\frac{4\delta}{7G}$~:
   \begin{align*}
     R(h, \C_j)=\EE_{(\obs,y)\sim\D} [\mu_h(\obs) &\neq y \wedge \obs \in \C_j] \le ~ {\widehat R}_{\rho}(h, \C_j) + \frac{\eta}{G} + \frac{2\kappa}{\rho}\fR_{n,j}^*(\F_\HHH) + \frac{2K}{\rho} \fR^*_{u,j}(\F_\HHH) \\  &~   + 5\sqrt{\frac{\kappa \neta(j)\log \frac{14G}{\delta}}{2n^2}} + 5\sqrt{\frac{\kappa \ueta(j)\log \frac{14G}{\delta}}{2u^2}} + \frac{7\log \frac{14G}{\delta}}{3(n-1)}  + \frac{7\log \frac{14G}{\delta}}{3(u-1)},
   \end{align*}
\noindent
where $\neta(j) = |\lda \cap \C_j|$,  and $\fR_{n,j}^*(\F) = \EE_{\sigma,\lda} \sup\limits_{f\in\F_\HHH} \frac{2}{n}\biggl|\sum_{\obs_i\in\lda\cap\C_j}\sigma_i f(\obs_i)\biggr|$, and $\ueta(j) = |\ulda \cap \C_j|$, and $\fR_{u,j}^*(\F) = \EE_{\sigma,\ulda} \sup\limits_{f\in\F_\HHH} \frac{2}{u}\biggl|\sum_{\obs_i\in\ulda\cap\C_j}\sigma_i f(\obs_i)\biggr|.$ Summing up over all clusters it comes
\begin{align*}
     \sum\limits_{\C_j\in \C_\kappa(\eta)} R(h, \C_j) \leq & \sum\limits_{\C_j\in \C_\kappa(\eta)}{\widehat R}_{\rho}(h, \C_j) + \eta + \frac{2\kappa}{\rho}\fR_{n}^*(\F_\HHH) +  \frac{2K}{\rho} \fR^*_{u}(\F_\HHH)  +\\  & 5\sum_{j=1}^G\sqrt{\frac{\kappa \neta(j)\log \frac{14G}{\delta}}{2n^2}} + 5\sum_{j=1}^G\sqrt{\frac{\kappa \ueta(j)\log \frac{14G}{\delta}}{2u^2}} + \frac{7G\log \frac{14G}{\delta}}{3(n-1)}+\frac{7G\log \frac{14G}{\delta}}{3(u-1)}.
\end{align*}

By the Cauchy--Schwarz inequality $(\sum_{i=1}^G a_i b_i)^2 \le (\sum_{i=1}^G a_i^2)  (\sum_{i=1}^G b_i^2)$, then by fixing $b_i = 1, \forall i\in\{1,\ldots,G\}$, we can bound the two last terms of the right hand side inequality, and get 
\begin{align}
\label{ineq:x02}
\sum\limits_{\C_j\in \C_\kappa(\eta)} R(h, \C_j) \leq &\sum\limits_{\C_j\in \C_\kappa(\eta)}{\widehat R}_{\rho}(h, \C_j) + \eta + \frac{2\kappa}{\rho}\fR_{n}^*(\F_\HHH) +  \frac{2K}{\rho} \fR^*_{u}(\F_\HHH) +  \\
   &~  5\sqrt{\frac{G \neta \kappa \log \frac{14KG}{\delta}}{2n^2}} + 5\sqrt{\frac{G \ueta \kappa \log \frac{14KG}{\delta}}{2u^2}}+\frac{7G\log \frac{14G}{\delta}}{3(n-1)}+\frac{7G\log \frac{14G}{\delta}}{3(u-1)},\nonumber
\end{align}
with $n^*_\eta=\sum_{\C_j\in \C_\kappa(\eta)}=n^*_\eta(j)$ and $u^*_\eta=\sum_{\C_j\in \C_\kappa(\eta)}=u^*_\eta(j)$.



%
From the inequality $\eta\leq \Delta_n(\A_\ulda,\A^\star,\lda)$ and Lemma 1, the following upper-bound holds with probability at least $1-\frac{\delta}{7}$~:
\[
\eta\leq \frac{L}{u} + L\sqrt{\frac{\log \frac{14}{\delta}}{2u}} + \sqrt{\frac{ \log \frac{14}{\delta}}{2n}}
\]
By the inequality $\forall a>0, b>0; (a+b)^2 \le 2(a^2 + b^2)$ it then comes~:
\begin{equation}
\label{eq:x04}
\eta\leq \frac{L}{u} + \sqrt{\left(\frac{L^2}{u} + \frac{ 1}{n}\right)\log \frac{14}{\delta}}
\end{equation}

Further the risk of classification outside the set of confident clusters can be decomposed as~:
\begin{align}
  \label{ineq:x03}
\EE_{(\obs,y)\sim\D}[\mu_h(\obs)\neq y\wedge \obs \in\lda\setminus\C_{\kappa}(\eta)] = ~& \EE_{(\obs,y)\sim\D}[\mu_h(\obs)\neq y\mid \obs \in\lda\setminus\C_{\kappa}(\eta)]\times \nonumber \\ 
& \EE_{(\obs,y)\sim\D}[\obs \in\lda\setminus\C_{\kappa}(\eta)].
\end{align}
 Similarly to the previous  development, and from the multi-class classification generalization bound and the Data-dependent Bennett's inequality (appendix A, theorems \ref{thrm:lei} and \ref{thrm:maurer}), the above risk is upper-bounded with probability at least $1-2\frac{\delta}{7}$ by~:

\begin{align}
  \label{eq:x05}
\EE_{(\obs,y)\sim\D}[\mu_h(\obs)\neq y\mid \obs \in\lda\setminus\C_{\kappa}(\eta)]\le &~ 
\frac{1}{n}\sum_{(\obs,y)\in\lda\setminus\C_\kappa(\eta)}\Phi_\rho(m_h(\obs,y))+\frac{2K}{\rho}\fR_n(\F_\HHH)+\nonumber\\
&~5\sqrt{\frac{K(n-\neta)\log\frac{14K}{\delta}}{2n^2}}+\frac{7\log\frac{14}{\delta}}{3(n-1)}
\end{align}

The result then follows from the inequalities, $\forall a>0, b>0, c>0; (a+b+c)^2 \le 3(a^2 + b^2+ c^2)$; $5\sqrt{3} < 9$;  \eqref{ineq:x01}, \eqref{ineq:x02}, \eqref{eq:x04}, \eqref{eq:x05} and the union-bound.    
\end{Proof}

\newpage

\vskip 0.2in
\bibliography{sample}
\bibliographystyle{theapa}

\end{document}